\documentclass[journal]{IEEEtai}

\usepackage[colorlinks,urlcolor=blue,linkcolor=blue,citecolor=blue]{hyperref}

\usepackage{color,array}

\usepackage{graphicx}

\usepackage{times}
\usepackage{epsfig}
\usepackage{graphicx}
\usepackage{amsmath}
\usepackage{amssymb}

\usepackage[dvipsnames]{xcolor}
\usepackage{booktabs}
\usepackage{multirow}
\usepackage{fixmath}
\usepackage{subcaption}
\usepackage{float}
\usepackage{xspace}
\usepackage{comment}
\usepackage{hyperref}
\usepackage[normalem]{ulem}
\usepackage[ruled]{algorithm2e}
\usepackage{wrapfig}

\newcommand{\added}[1]{#1}

\newcommand{\addedrev}[1]{#1}

\newcommand{\std}[1]{\scriptsize{$\pm$#1}}

\begin{document}
\title{Unsupervised Learning of Unbiased Visual Representations}

\author{Carlo Alberto Barbano, \IEEEmembership{Member, IEEE}, Enzo Tartaglione \IEEEmembership{Senior Member, IEEE}, Marco Grangetto, \IEEEmembership{Senior Member, IEEE}
\thanks{Submitted 9 Jan 2024. Accepted for publication 28 Aug 2024. %
C. A. Barbano is with the University of Turin, Italy (e-mail: carlo.barbano@unito.it). E. Tartaglione is with Télécom Paris, LTCI, IP Paris, France (enzo.tartaglione@telecom-paris.fr). M. Grangetto is with the University of Turin, Italy (e-mail: marco.grangetto@unito.it).}}

\markboth{Journal of IEEE Transactions on Artificial Intelligence, Vol. 00, No. 0, Month 2020}
{Carlo A. Barbano \MakeLowercase{\textit{et al.}}: Unsupervised Learning of Unbiased Visual Representations}
\maketitle

\begin{abstract}
Deep neural networks often struggle to learn robust representations in the presence of dataset biases, leading to suboptimal generalization on unbiased datasets. This limitation arises because the models heavily depend on peripheral and confounding factors, inadvertently acquired during training. Existing approaches to address this problem typically involve explicit supervision of bias attributes or reliance on prior knowledge about the biases.
In this study, we address the challenging scenario where no explicit annotations of bias are available, and there's no prior knowledge about its nature. We present a fully unsupervised debiasing framework with three key steps: firstly, leveraging the inherent tendency to learn malignant biases to acquire a bias-capturing model; next, employing a pseudo-labeling process to obtain bias labels; and finally, applying cutting-edge supervised debiasing techniques to achieve an unbiased model. Additionally, we introduce a theoretical framework for evaluating model biasedness and conduct a detailed analysis of how biases impact neural network training. Experimental results on both synthetic and real-world datasets demonstrate the effectiveness of our method, showcasing state-of-the-art performance in various settings, occasionally surpassing fully supervised debiasing approaches.
\end{abstract}

\begin{IEEEImpStatement}
This study addresses a critical challenge in deep neural networks by tackling the issue of biased representations without the need for explicit annotations or prior knowledge about the biases. The proposed unsupervised debiasing framework introduces a novel approach involving three steps: leveraging the network's tendency to learn biases, employing a pseudo-labeling process to acquire bias labels, and applying advanced supervised debiasing techniques. The research not only presents a practical solution for mitigating biases but also introduces a theoretical framework to evaluate model biasedness, providing a comprehensive understanding of the impact of biases on neural network training. The experimental results, conducted on both synthetic and real-world datasets, demonstrate the method's effectiveness, showcasing state-of-the-art performance in various settings and occasionally surpassing fully supervised debiasing approaches. This work is poised to significantly contribute to the enhancement of model generalization and reliability in the face of dataset biases, advancing the field of deep learning towards more robust and unbiased artificial intelligence systems.
\end{IEEEImpStatement}

\begin{IEEEkeywords}
Bias, Deep learning, Regularization, Unsupervised learning.
\end{IEEEkeywords}

\section{Introduction}
\label{sec:introduction}
\IEEEPARstart{D}{eep} learning has reached state-of-the-art performance in a large variety of tasks, for example in computer vision and natural language processing~\cite{voulodimos2018deep, NEURIPS2020_1457c0d6}.
Furthermore, the broad availability of computational resources allowed researchers to train increasingly larger models~\cite{dosovitskiy2020vit, NEURIPS2020_1457c0d6}, resulting in better generalization capabilities. 
Deep learning-based tools are impacting more and more everyday life and decision-making~\cite {laumer2015impact}. This is one of the reasons why AI trustworthiness has been recognized as a major prerequisite for its use in society~\cite{hleg2019ethics, zhang2019artificial}. 
Following the guidelines of the European Commission of 2019, these tools should be \emph{lawful}, \emph{ethical} and \emph{robust}.
Providing a warranty on this topic is currently a matter of study and discussion~\cite{schramowski2020making, stock2020eccv, teso2019aies_XIML, revisetool, barbano2021bridging}.

In recent years, it has also become apparent how deep neural networks often rely on erroneous correlations and biased features present in the training data. Many works have proposed solutions for this matter, for example addressing gender biases when dealing with facial images~\cite{nam2020learning, Kim_2019_CVPR, zhao2021learning, lee2021learning} or with medical data~\cite{tartaglione2021end, dufumier2021brain}. 
When dealing with biases, many of the proposed techniques often work in a \emph{supervised} manner, assuming that each sample can be labeled with a corresponding bias class~\cite{alvi2018turning, tartaglione2021end,Kim_2019_CVPR}.
Other works like \cite{bahng2019rebias} and \cite{cadene2019rubi} on the other hand, only make assumptions about the nature of bias, for example, for designing the model architecture. Compared to the fully supervised setting, this is a more realistic scenario, however, it still requires some form of prior knowledge about the bias.

In this work, we propose an \emph{unsupervised} debiasing strategy which allows us to obtain an unbiased model when the data contain biases that are either unknown or not labeled. \added{We also show how this formulation can be effective in presence of multiple biases, such as on facial images.} 
We present a framework for retrieving information about bias classes in the data, in a completely unsupervised manner. First, we show how to exploit the naturally biased representations to retrieve the presence of unknown biases. 
Then, by performing a pseudo-labeling step, we are able to employ traditional supervised debiasing techniques for learning an unbiased model. For this purpose, we focus on the recently proposed EnD technique, proposed by \cite{tartaglione2021end}, which obtains state-of-the-art results in different debiasing tasks, and we name the unsupervised extension~\emph{U-EnD} (Unsupervised EnD), although our framework is general and can be adapted to different supervised techniques with minimal effort. 
We also study the effect of the strength of biases on the learning process, by developing a mathematical framework that allows us to determine how \emph{biased} a model is, backed by empirical experimentation, on controlled cases.
Our results also further back up the findings about how neural networks tend to prefer simpler and easier patterns during the learning process, as also confirmed in \cite{arpit2017memorization} and \cite{nam2020learning}.
In a summary, with U-EnD:
\begin{enumerate}
    \item we propose an unsupervised framework for learning unbiased representations, with no need for explicit supervision on the bias features;
    \item we propose a simple but effective mathematical model that can be used to quantitatively measure how much a model is biased;
    \item we validate our proposed framework and model on state-of-the-art synthetic and real-world datasets.
\end{enumerate}

The rest of the work is structured as follows: In Section~\ref{sec:related} we review the literature related to debiasing in deep neural networks; in Section~\ref{sec:analysis-mnist} we propose a preliminary experiment where we analyze how bias is captured by an un-regularized model, leading to the proposal of a formulation for model biasedness; in Section~\ref{sec:unsupervised-end} we present the extension for our proposed unsupervised framework for debiasing; finally, in Section~\ref{sec:experiments} we provide empirical results on real-world data and in Section~\ref{sec:conclusion}, the conclusions are drawn.

\section{Related works}
\label{sec:related}

Addressing the issue of biased data and how it affects neural network generalization has been the subject of numerous works.
Back in 2011, Torralba \emph{et al.}~\cite{torralba2011unbiased} showed that many of the most commonly used datasets are affected by biases. In their work, they evaluate the cross-dataset generalization capabilities based on different criteria, showing how data collection could be improved. \added{\cite{Khosla2012UndoingTD} employs max-margin learning (SVM) to explicitly model dataset bias for different vision datasets.}
With a similar goal, \cite{tommasi2017deeper} propose different benchmarks for cross-dataset analysis, aimed at verifying how different debiasing methods affect the final performances.
Data collection should be carried out with great care, in order not to include unwanted biases. Leveraging data already publicly available could be another way of tackling the issue. \cite{gupta2018robot} explore the possibility of reducing biases by exploiting different data sources, in the practical context of sensors-collected data. They propose a strategy to minimize the effects of imperfect execution and calibration errors, showing improvements in the generalization capability of the final model. 
Another issue related to the unwanted learning of spurious correlations in the data has been highlighted by recent works. For example, \cite{song2017machine} and \cite{barbano2021bridging} show how traditional training approaches allow information not relevant to the learning task to be stored inside the network. The experiments carried out in these works show how accurately some side information can be recovered, resulting in a potential lack of privacy.
\cite{beutel2019putting} provide insights on algorithmic fairness in a production setting, and propose a metric named \emph{conditional equality}. They also propose a method, absolute correlation regularization, for optimizing this metric during training.
Another possibility of addressing these issues on a data level is to employ generative models, such as GANs, to clean up the dataset with the aim of providing fairness: some examples lie in the approach by \cite{xu2018fairgan} and \cite{sattigeri2018fairness}.
\cite{madras2018learning} also employs a GAN to obtain fair representations. Following up on data synthesizing approaches, \cite{lee2021learning} introduces DFA, a novel feature-level data augmentation technique that leverages a disentangled representation of intrinsic and bias attributes. DFA enables the synthesis of diverse bias-conflicting samples, crucial for effective debiasing and improved generalization in image classification models, outperforming existing methods on synthetic and real-world datasets.

All of the above-mentioned approaches generally deal directly with the data level and provide useful insights for designing more effective debiasing techniques. 
In the related literature, we can most often find debiasing approaches based on ensembling methods, adversarial setups, or regularization terms which aim at obtaining an \emph{unbiased} model using \emph{biased} data. We distinguish three different classes of approaches, in order of complexity: those that need full explicit supervision on the bias features (e.g. using bias labels), those that do not need explicit bias labels but leverage some prior knowledge of the bias features, those which no dot need neither supervision nor prior-knowledge.

\subsection{Supervised approaches}
\label{sec:related-supervised}
Among the relevant related works, the most common debiasing techniques are supervised, meaning that they require explicit bias knowledge in the form of labels. The most common approach is to use an additional bias-capturing model, with the task of specifically capturing bias features. This bias-capturing model is then leveraged, either in an adversarial or collaborative fashion, to enforce the selection of unbiased features on the main model. 
\added{We can find the typical supervised adversarial approach in \cite{alvi2018turning,Xie2017ControllableIT}.
Specifically, \cite{Xie2017ControllableIT} proposes an adversarial framework for learning invariant representations with respect to some attributes in the data. In \cite{alvi2018turning}, they employ an explicit bias classifier, which is trained on the same representation space as the target classifier, using a min-max optimization approach. In this way, the shared encoder is forced to extract unbiased representations.}
Similarly, \cite{Kim_2019_CVPR} proposes Learning Not to Learn (LNL), which leverages adversarial learning and gradient inversion to reach the same goal.
Adversarial approaches can be found in many other works, for example in the work by \cite{wang2019iccv}, where they show that biases can be learned even when using balanced datasets, and they adopt an adversarial approach to remove unwanted features from intermediate representations of a neural network. 

\added{Moving away from adversarial approaches, 
the authors in~\cite{ClarkYZ19} propose LearnedMixin: they train a biased model with explicit supervision on the bias labels, and then they build a robust model forcing its prediction to be made on different features. Wang \emph{et al.}~\cite{wang2020fair} perform a thorough review of the related literature and propose a technique based on an ensemble of classifiers trained on a shared feature space.}

Another possibility is represented by the application of adjusted loss functions or regularization terms.
For example, \cite{sagawa2019distributionally} propose Group-DRO, which aims at improving the model performance on the \emph{worst-group} in the training set, defined based on prior knowledge of the bias distribution. 
\added{\cite{tartaglione2021end} proposes EnD (see Sec.~\ref{sec:supervised-end}), which consists in a regularization term to learn bias-invariant representations.}
\addedrev{Another approach is represented by FairKL~\cite{barbano2023unbiased}, where a regularization term is employed to minimize the KL-diverge between bias-conflicting and bias-aligned samples.}

\subsection{Prior-guided approaches}
In many real-world cases, explicit bias labels are not available. However, it might still be possible to make some assumptions or have some prior knowledge about the nature of the bias.

\added{Borrowing from domain generalization techniques, another kind of approach aimed at learning robust representation is proposed by \cite{wang2018learning} with HEX. They propose a differentiable neural-network-based gray-level co-occurrence matrix, inspired by~\cite{haralick1973textural} and \cite{lam1996glcm}, to extract biased textural information, which is then employed for learning invariant representations.
A different context is presented by \cite{hendricks2018women}. They propose an equalizer model and a loss formulation that explicitly takes into account gender bias in image captioning models. In this work, the prior is given by annotation masks indicating which features in an image are appropriate for determining gender. Related to this approach, another possibility is to constrain the model prediction to match some prior annotation of the input, as done in~\cite{Ross2017RightFT}, where gradients re-weighting is used to encourage the model to focus on the right input regions. Similarly, \cite{Selvaraju_2019_ICCV} propose HINT, which optimizes the alignment between account manual visual annotation and gradient-based importance masks, such as Grad-CAM, proposed by~\cite{selvaraju2017grad}.}

Similarly to the work presented in Sec.~\ref{sec:related-supervised}, Banhg. \emph{et al.}~\cite{bahng2019rebias} propose ReBias, an ensembling-based technique. They build a bias-capturing model (an ensemble in this case). The prior knowledge about the bias is used in designing the bias-capturing architecture (e.g. by using a smaller receptive field for texture and color biases). The optimization process consists of solving a min-max problem to promote independence between the biased representations and the unbiased ones.
A similar assumption for building the bias-capturing model is made by \cite{cadene2019rubi} with RUBi. In this work, logit re-weighting is used to promote independence of the predictions on the bias features.
Regarding samples re-weighting, \cite{liu2021just} introduce JTT, a two-stage training approach for mitigating low accuracy on specific groups caused by spurious correlations in standard empirical risk minimization (ERM) models. By up-weighting misclassified examples from underperforming groups, JTT significantly narrows the worst-group accuracy gap between standard ERM and group distributionally robust optimization (group DRO) across diverse image classification and natural language processing tasks, with minimal reliance on group annotations. Following on similar ideas, \cite{hong2021unbiased} address dataset bias in machine learning training by introducing the Bias-Contrastive (BiasCon) loss, utilizing contrastive learning within a bias-aware framework to improve model generalization across shifted correlations. The proposed Bias-Balanced (BiasBal) regression and Soft Bias-Contrastive (SoftCon) loss further enhance debiasing performance, demonstrating significant improvements over prior methods across diverse realistic datasets.

\subsection{Unsupervised approaches}
Increasing in complexity, we consider as unsupervised approaches those methods which do not \emph{i)} require explicit bias information \emph{ii)} use prior knowledge to design specific architectures. 
In this setting, building a bias-capturing model is a more difficult task, as it should rely on more general assumptions.
\added{Ji \emph{et al.}~\cite{ji2019invariant} propose an unsupervised clustering method that can learn representations invariant to some unknown or ``distractor'' classes in the data, by employing over-clustering. Although not strictly for debiasing purposes, another clustering-based technique is proposed by~\cite{van2020scan}: they employ a two-step approach for unsupervised learning of representations, where they mine the dataset to obtain pseudo-labels based on neighbor clusters. This approach is also closely related to our work. A very similar and recently proposed approach can be found in~\cite{nam2022ssa} with SSA: similarly to this work, the authors propose to assign pseudo-labels based on biased clusters. However, different from our proposed method, they still require a smaller set with bias annotations.}
Nam \emph{et al.}~\cite{nam2020learning} proposes a technique named Learning from Failure (LfF). They exploit the training dynamics: a bias-capturing model is trained with a focus on \emph{easier} samples, using the Generalized Cross-Entropy (GCE) loss proposed by~\cite{zhang2018gce}, which is assumed to be aligned with the bias, while a debiased network is trained emphasizing samples which the bias-capturing model struggles to learn. These assumptions they make are especially relevant to our work, as U-EnD also leverages the training dynamics for building the bias-capturing model.
Similar assumptions are also made by~\cite{luo2022pseudo} where GCE is also used for dealing with biases in a medical setting using Chest X-ray images. \addedrev{More recently, Zhang~\emph{et~al}~\cite{zhang2023learning} propose an unsupervised approach that leverages samples near the decision boundary (i.e. bias-aligned or bias-conflicting samples) to correct the model's prediction.}
\addedrev{In~\cite{zhang2024poisoning}, authors propose a data poisoning approach to ultimately emphasize the weights of bias-conflicting samples, to prevent the model from learning spurious correlations.}
\addedrev{Another approach is represented by~\cite{Wang_2024_CVPR}, where a method based on Neural Collapse~\cite{papyan2020prevalence} is used to prevent models from learning simple shortcuts, focusing more on true correlations in the data.}

\section{\added{The EnD regularization}}
\label{sec:supervised-end}

\begin{table}
   \center
    \caption{\added{Overview on the notation used in this work.}}
    \begin{tabular}{c l}%
        \toprule
        \textbf{Symbol} & \textbf{Meaning}\\
        \midrule
        $x_i$      & input ($i$-th sample in the dataset)\\
        $y_i$      & predicted class for the $i$-th sample \\
        $t_i$      & target class for the $i$-th sample \\
        $b_i$      & bias class for the $i$-th sample \\
        
        $Y$        & random variable associated to the predicted labels\\
        $T$        & random variable associated to the target labels\\
        $B$        & random variable associated to the bias labels\\
        $N_T$                     & cardinality of targets\\

        \bottomrule
    \end{tabular}
    \label{tab:notationover}
\end{table}

\added{In this section, we introduce the notation we adopt in this work and provide a detailed explanation of how the EnD~\cite{tartaglione2021end} debiasing technique works in a supervised setting. Then, in Sec.~\ref{sec:unsupervised-end} we will move on to the unsupervised extension. 
The notation introduced in this work is summarized in Table~\ref{tab:notationover}\footnote{\added{For the rest of this work, we conform to the standard notation proposed by Goodfellow~\emph{et~al.}~\cite{GoodBengCour16}, available at \url{https://github.com/goodfeli/dlbook\_notation}}}.}
\added{Given a neural network \emph{encoder} $f(\cdot) \in R^N$ 
which extracts feature vectors of size $N$ and a \emph{classifier} $g(\cdot) \in R^{N_T}$ which provides the final prediction, we consider the neural network $(g~\circ~\gamma~\circ~f)(\cdot)$
where $\gamma : \mathbb{R}^N \rightarrow \mathbb{R}^N$ is a normalization function to obtain $z = \gamma(x) = x/\|x\|_2$.
The EnD regularization term $\mathcal{R}$ is applied jointly with the loss function $\mathcal{L}$ (e.g. cross-entropy), forcing $(\gamma \circ f)(\cdot)$ to filter out biased features from the extracted representation $z$. 
Hence, the overall objective function we aim to minimize is:}
\begin{equation}
    \label{eq:uprule}
    \mathcal{J} = \mathcal{L} + \mathcal{R},
\end{equation}
\added{where $\mathcal{R}$ is the sum of the disentangling and entangling terms, weighted by two hyper-parameters $\alpha$ and $\beta$:}
\begin{equation}
\label{eq:objective-func}
    \mathcal{R} = \alpha\mathcal{R}^{\perp} + \beta\mathcal{R}^{\parallel}
\end{equation}
\added{Within a minibatch, let $i \in I \equiv \{1\dots M\}$ be the index of an arbitrary sample. We define $y_i$, $t_i$, and $b_i$ as the predicted, ground truth target and bias classes, respectively. 
The disentangling term $R^{\perp}$ is defined, for the $i$-th sample, as:}
\begin{equation}
	\mathcal{R}^{\perp}_{i} = \frac{1}{|B(i)|} \sum_{a \in B(i)} | z_i \cdot z_a |
\end{equation}
\added{where $B(i) := \{j \in I \mid b_j = b_i  \} \setminus \{i\}$ is the set of all samples sharing the same bias class of $x_i$.} 

\added{The goal of this term is to suppress the common features among samples that have the same bias. %
The entangling term $R^{\parallel}$ is defined, for the $i$-th sample, as:}
\begin{equation}
	\mathcal{R}^{\parallel}_{i} = - \frac{1}{|J(i)|}\sum_{j \in J(i)} z_i \cdot z_j
\end{equation}
\added{where $J(i) := \{ j \in I \mid t_j = t_i \} \setminus B(i)$ is the set of all samples sharing the same target class of $x_i$ but with different biases. %
Complementarily to the disentangling term, the goal of this term is to encourage correlation among samples of the same class but with different biases, in order to introduce invariance with respect to the biased features.
So, for the $i$-th sample, the entire EnD regularization term $\mathcal{R}_i$ can be written as:}
\begin{equation}
\label{eq:end-full-new}
  \mathcal{R}_i =  \alpha\frac{1}{|B(i)|} \sum_{a \in B(i)} | z_i \cdot z_a | - \beta \frac{1}{|J(i)|}\sum_{j \in J(i)} z_i \cdot z_j \,. 
\end{equation}
\added{The final $\mathcal{R}$ of Eq.~\ref{eq:objective-func}, is then just computed as the average over the minibatch:}

\begin{equation}
    \mathcal{R} = \frac{1}{M}\sum_i \mathcal{R}_i
\end{equation}
\begin{figure}
    \centering
    \includegraphics[width=0.6\columnwidth,trim={110 10 140 0},clip]{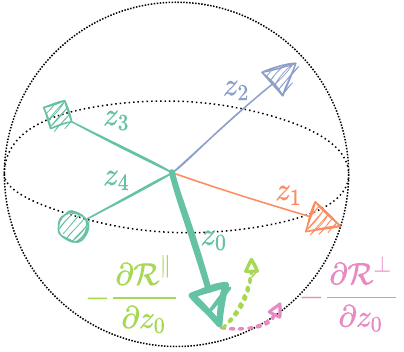}
    \caption{\added{Effect of the regularization term \eqref{eq:end-full-new} with respect to $z_0$. The features extracted from samples belonging to the same target class (same arrow's vertex) are entangled through the $R^\parallel$ term (in light green) while features for the same bias class (same color in this case, with respect to $z_0$, dark green) are disentangled through the $R^\perp$ term (in pink).}}
    \label{fig:notation}
\end{figure}

\added{To visualize the effect of $\mathcal{R}$ as expressed in Eq.~\ref{eq:end-full-new}, consider a simple classification problem with three target classes and three different biases as illustrated in Fig.~\ref{fig:notation}. 
Training a model without explicitly addressing the presence of biases in the data, will most likely result in representations aligned by the bias attributes rather than the actual target class (Fig.~\ref{fig:notation}). 
The goal of $\mathcal{R}$ is to encourage the alignment of representations based on the correct features by \emph{i.)} disentangling representations of the same bias ($\mathcal{R}^\perp$) and \emph{ii.)} entangling representations of the same target in order to introduce invariance to the bias features ($\mathcal{R}^\parallel$).}

\section{Unsupervised debiasing by subgroup discovery}
\label{sec:unsupervised-end}

\begin{figure*}
    \centering
    \includegraphics[width=\textwidth]{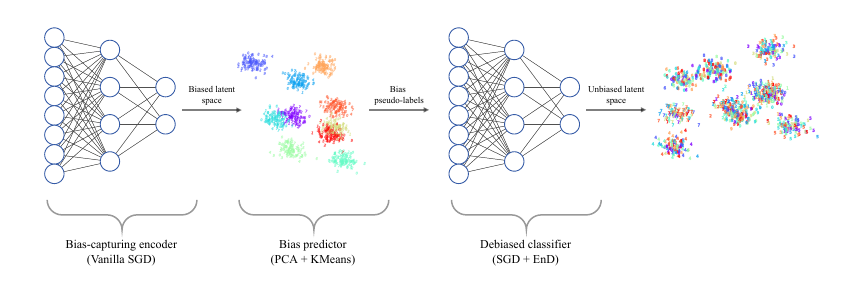}
    \caption{Overview of our unsupervised debiasing approach: first we train a bias-capturing encoder, then we determine bias pseudo-labels with a bias predictor. Finally, we employ the predicted labels for training a final debiased classifier. In this figure, we use Biased MNIST~(\cite{bahng2019rebias}) as an example, where the bias is given by a strong correlation between digit and color.}
    \label{fig:overview-unsupervised}
\end{figure*}

\begin{algorithm}[t]
\DontPrintSemicolon
\SetKwBlock{stepA}{Train bias-capturing model}{end}
\SetKwBlock{stepB}{Train bias predictor}{end}
\SetKwBlock{stepC}{Train unbiased classifier}{end}
\KwIn{
    \begin{minipage}{.75\linewidth}
    Training and validation data $X^t = \{(x_i, y_i)\}$, $X^v = \{(\hat{x}_i, \hat{y}_i)\}$;
    Randomly initialized parameters $\theta_B = \{\theta_f, \theta_g\}$ and $\theta_D = \{\theta_f^D, \theta_g^D\}$ of the biased and unbiased classifiers.
    \end{minipage}
}
\BlankLine

\KwOut{Trained parameters $\theta_D$ of the unbiased classifier.}
\BlankLine

\stepA{
\par
Train the biased classifier using vanilla SGD: $\theta_B \leftarrow \text{SGD}(\theta_B, X^t)$\;
Compute the biased representations:
$Z^t = \{ f(x; \theta_f) \} \quad \forall x \in X^t$ and
$Z^v = \{ f(x; \theta_f) \}\quad\forall x \in X^v$\;
}

\stepB{
Compute the PCA projections $P^t, P^v$ of $Z^t,Z^v$\; 
Fit $k$ clusters on $P^v$ choosing the optimal $k$ based on silhoutte and compute the cluster centroids:
$\{\mu_1,...,\mu_k\} \leftarrow \text{KMeans}(P^v, k)$\;
Assign the pseudo-labels $\hat{b}_i \leftarrow \underset{b \leq k}{\arg\!\min}\,(P^t_i,\, \mu_b)$\; 
Update the training set $X^t \leftarrow \{(x_i, y_i, \hat{b}_i)\}$\;
}

\stepC{
Learn the parameters $\theta_D$ on $X^t$ searching the optimal $\alpha$ and $\beta$ on $X^v$:
$\theta_D \leftarrow \text{SGD}(\theta_D, X^t) + R(\theta^D_f, X^t, \alpha, \beta)$\;
}

\caption{General scheme of U-EnD}
\label{alg:overview}
\end{algorithm}

\noindent In this section we present our proposed unsupervised end-to-end debiasing approach, showing how an explicitly supervised technique such as EnD can be extended to the unsupervised case, where the bias labels are unavailable. 
We do this by showing how the information of the bias can be partially, and sometimes fully, recovered in a completely unsupervised manner. 
To achieve that, our proposed algorithm consists of three sequential steps, as illustrated in Fig.~\ref{fig:overview-unsupervised}. \added{The steps can be summarized as follows.}

\begin{enumerate}
    \item \added{First, we train a bias-capturing classifier, employing standard optimization techniques, e.g. SGD or Adam (Fig.~\ref{fig:overview-unsupervised} left).}
    \item \added{Then, we recover bias-related information from the latent space of the biased classifier via clustering, to obtain a bias predictor, which we employ to categorize all of the training samples into different bias classes (Fig.~\ref{fig:overview-unsupervised} center).}
    \item \added{Lastly, we apply the EnD debiasing technique using the predicted bias labels, to obtain a debiased classifier (Fig~\ref{fig:overview-unsupervised} right).}
\end{enumerate}
 
A general scheme of the entire pipeline can be found in Algorithm~\ref{alg:overview}. 
Throughout this work, we assume that an \emph{unbiased} validation set is available: this is needed for searching the optimal EnD hyper-parameters.

\subsection{Training a bias-capturing model}
\label{sec:phase1}
The first step of our proposed algorithm is to train a bias-capturing model, which in our case is represented by a \emph{biased} encoder. To achieve this, we perform a vanilla training of a CNN classifier on the available training data. Here, we do not employ any technique aimed at dealing with the presence of biases in the data. The intuition of this approach is that if bias features are easier to learn than the desired target attributed, then the resulting model will also be biased (\added{see Sec.~\ref{sec:analysis-mnist}}). This has also been observed in other works. Following the definition proposed by Nam~\emph{et al.}~\cite{nam2020learning} we can identify two cases:

\begin{itemize}
    \item \emph{benign biases:} even if biases are present in the data, the model does not rely on bias-related features as the target task is easier to learn;
    \item \emph{malignant biases:} biases are present in the data, and they are easier for the model to learn instead of the target task.
\end{itemize}

\noindent The latter case is the most relevant for our work, as malignant biases are those that result in a loss of performance when evaluating the model on an unbiased set.
Fig.~\ref{fig:overview-unsupervised} shows a visualization of the embeddings obtained with a biased encoder on the BiasedMNIST~\cite{bahng2019rebias} dataset, where the background color correlates very well with the target digit class, as shown in Fig.~\ref{fig:biased-mnist}.
It is clear how the different clusters emerging in the latent space correspond to the different background color, rather than to the actual digit.
This first step is summarized in Algorithm~\ref{alg:overview}, and we now provide a more formal description.
Let $\theta_B = \{\theta_f, \theta_g\}$ be the set of parameters of the bias-capturing model $p(x; \theta) = g(f(x; \theta_f); \theta_g)$ where $f$ and $g$ are the encoder and the classifier, respectively. The objective function we aim to minimize is the cross-entropy loss (CE):
\begin{equation}
    \label{eq:objective-fun-biased}
    \mathcal{L}_\text{CE}(p(x; \theta_B), q(x)) = -\sum_{t \in T} q(t|x) \log p(t|x;\theta_B),
\end{equation}
where $q(x)$ represents the ground truth class distribution. We say that there is a benign bias in the dataset, if we can identify some distribution $r(x)$, related to some other confounding factor in the data, such that there exists a set of parameters $\theta'$ which is a local minimizer of~\eqref{eq:objective-fun-biased} and
$\theta' = \arg\!\min_{\theta_B} \mathcal{L}_\text{CE}(p(x; \theta_B), r(x))$. 
If, additionally, $r(x)$ is also easier to approximate than $q(x)$, then the bias is malignant and by applying the optimization process we obtain a bias-capturing model.
Once the biased model is trained, we only consider the encoder $f(x; \theta_f)$, as we are interested in analyzing its latent space to retrieve bias-related information. 

\subsection{Fitting a bias predictor}
\label{sec:phase2}

The second step consists in obtaining a predictor which can identify the bias in the data. 
Based on the observations made in Section~\ref{sec:phase1}, we employ a clustering algorithm to categorize the extracted representations into different classes. As shown in Fig.~\ref{fig:overview-unsupervised}, the identified clusters correspond to the biases in the dataset.
In this work, we choose KMeans~\cite{lloyd1982least} as it is one of the most well-known clustering algorithms.
Given a set of representation $z = \{z_1, z_2,\dots, z_n\}$ extracted by $f(x; \theta_B)$ we aim to partition $z$ into $k$ sets $C = \{C_1, C_2, \dots, C_k\}$ to minimize the within-clusters sum of squares (WCSS), which can be interpreted as the distance of each sample from its corresponding cluster centroid, by finding: 
\begin{equation}
    \label{eq:kmeans-objective}
    \underset{C}{\arg\!\min} \sum_{i=1}^k \sum_{z \in C_i}\| z - \mu_i \|^2,
\end{equation}
where $\mu_i$ is the centroid (average) of $C_i$.
Furthermore, once the clusters have been determined, 
it is very easy to use the determined centroids for classifying a new sample $\hat{z}$ based on its distance, simply by finding 
\begin{equation}
    \label{eq:kmeans-predict}
    \hat{b} = \underset{i \leq k}{\arg\!\min} \|\hat{z} - \mu_i\|^2,
\end{equation}
where $\hat{b}$ denotes the resulting pseudo-label. 
The KMeans algorithm requires a pre-specified number of clusters $k$: in this work, we automatically tune this parameter based on the
best silhouette score~\cite{rousseeuw87silhouetteCluster}, obtained by performing a grid search in the range $[2, 15]$. 
Considering that the representations obtained on the training set might be over-fitted, we choose to minimize \eqref{eq:kmeans-predict} on the validation set. Then, once the centroids of the clusters have been found, we use them for pseudo-labeling the training set.
Additionally, as KMeans is based on Euclidean distance, which can yield poor results in highly dimensional spaces, we perform a PCA projection of the latent space before solving~\eqref{eq:kmeans-objective}~and~\eqref{eq:kmeans-predict}. For the same reasons as above, the PCA transformation matrix is also computed on the validation set.
We refer to the ensemble of the PCA+KMeans as \emph{bias predictor} model.
The cluster information is then used as a bias pseudo-label, as explained in Section~\ref{sec:phase3}. 

\subsection{Training an unbiased classifier}
\label{sec:phase3}
The third and final step of our proposed framework consists of training an unbiased classifier. For this purpose, we use the clusters discovered in the previous phase as pseudo-labels for the bias classes, as shown in Figure~\ref{fig:overview-unsupervised}. This allows us to employ the fully supervised EnD regularization term for debiasing, \added{presented in Sec.~\ref{sec:supervised-end}.} Here we follow the approach as in \cite{tartaglione2021end}.
Denoting with $\theta_D = \{\theta_f^D, \theta_g^D\}$ the parameters of the encoder and the classifier of the debiased model $~{p'(x; \theta_D) = g(\gamma(f(x; \theta_f^D)); \theta_g^D)}$.
The objective function that we aim to minimize in this phase is:
\begin{equation}
    \label{eq:objective-fun-unbiased}
    \mathcal{L}_\text{CE}(p'(x; \theta_D), q(x)) + R(\gamma(f(x; \theta_f^D)), q(x), b(x)),
\end{equation}
where $b(x)$ is the distribution corresponding to the pseudo-labels computed in the clustering step of Section~\ref{sec:phase2}. The closer $b(x)$ is to the real distribution $r(x)$, the more minimizing~\eqref{eq:objective-fun-unbiased} will lead to minimizing $R$ with respect to the unknown ground-truth bias labels.

\section{Analysis on controlled experiments}
\label{sec:analysis-mnist}

\begin{figure}
    \centering
    \includegraphics[width=\columnwidth]{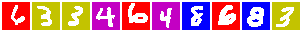}
    \caption{Biased MNIST by \cite{bahng2019rebias}. The bias is given by the correlation between digit and background color.}
    \label{fig:biased-mnist}
\end{figure}

In this section, we provide more insights supporting the proposed solution. 
To aid in the explanation of the different steps of the algorithm, we employ the Biased-MNIST dataset~\cite{bahng2019rebias} as a case study throughout this section. Biased-MNIST is built upon the original MNIST~\cite{lecun2010mnist} by injecting a color bias into the image's background, as shown in Fig.~\ref{fig:biased-mnist}. 
A set of ten default colors associated with each image is determined beforehand. 
To assign a color to an image, the default color is chosen with a probability $\rho$, while any other color is chosen with a probability of $(1-\rho)$. 
More explicitly, given a label $i, 0 \leq i < N_T$ with $N_T=10$,
the probability distribution for each color is given by: 
\begin{equation}
    \begin{cases}
        P(B = i|T = i) = \rho \\
        P(B \neq i | T = i) = \frac{1}{N_T - 1} (1 - \rho)
    \end{cases}
    \label{eq:dist-b-given-t}
\end{equation}
where $B$ and $T$ are random variables associated with the bias and target class respectively.
The $\rho$ parameter allows us to determine the degree of correlation between background color (\emph{bias}) and digit class (\emph{target}). %
Hence, higher values of $\rho$ correspond to more difficult settings.
Following~\cite{bahng2019rebias}, we generate different biased training set, by selecting $~{\rho \in \{0.990, 0.995, 0.997, 0.999 \}}$. To assess the generalization performance when training on a biased dataset, we construct an \emph{unbiased} test set generated with $\rho =\frac{1}{N_T} = 0.1$. Given the low correlation between color and digit class in the unbiased test set, models must learn to classify shapes instead of colors to reach a high accuracy. 

\begin{table}
    \centering
    \resizebox{\columnwidth}{!}{%
    \begin{tabular}{l c c c c}
        \toprule
        \multirow{2}{*}{Method} & \multicolumn{4}{c}{$\rho$ values}\\
                                & 0.999 & 0.997 & 0.995 & 0.990 \\ 
        \toprule
        Vanilla & 10.40\std{0.50} & 33.40\std{12.21} & 72.10\std{1.90} & \underline{89.10}\std{0.10} \\
        \added{LearnedMixin}~\cite{ClarkYZ19} & \underline{12.10}\std{0.80} & \underline{50.20}\std{4.50} & \underline{78.20}\std{0.70} & 88.30\std{0.70} \\
        EnD~\cite{tartaglione2021end} & \textbf{52.30}\std{2.39}    & \textbf{83.70}\std{1.03}  & \textbf{93.92}\std{0.35}  & \textbf{96.02}\std{0.08} \\
        \midrule
        \added{HEX}$^\dagger$~\cite{wang2018learning} & 10.80\std{0.40} & 16.60\std{0.80} & 19.70\std{1.90} & 24.70\std{1.60} \\
        \added{RUBi}$^\dagger$~\cite{cadene2019rubi} & 13.70\std{0.70} & 43.00\std{1.10} & \underline{90.40}\std{0.40} & \underline{93.60}\std{0.40}\\
        \added{ReBias}$^\dagger$~\cite{bahng2019rebias} & 22.70\std{0.40} & 64.20\std{0.80} & 76.00\std{0.60} & 88.10\std{0.60} \\
        U-EnD ($T$=80) & \underline{53.90}\std{4.03} & \underline{82.16}\std{0.63} & 74.39\std{0.43} & 88.05\std{0.16} \\
        U-EnD ($T$=10) & \textbf{55.29}\std{1.27} & \textbf{85.94}\std{0.33} & \textbf{92.92}\std{0.35} & \textbf{93.48}\std{0.06} \\
        \bottomrule
    \end{tabular}
    }
    \caption{Biased-MNIST accuracy on the unbiased test set. $T$ indicates the training iteration where the bias information is extracted.}
    \label{table:mnist-results}
\end{table}

\begin{figure*}
    \begin{subfigure}{0.33\textwidth}
        \centering
        \includegraphics[width=\columnwidth]{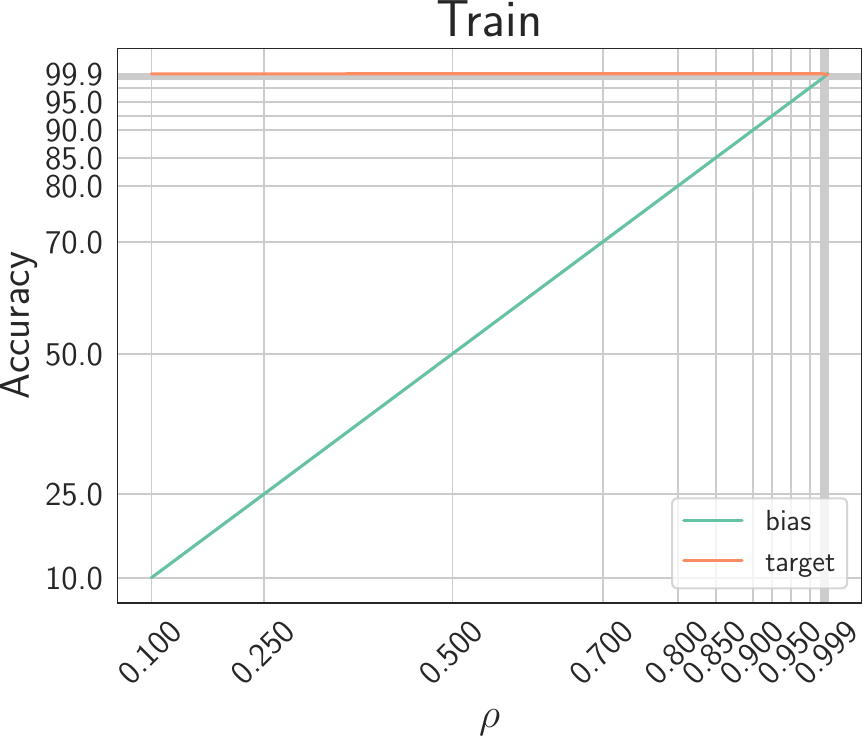}
        \caption{~}
        \label{fig:mnist-train}
    \end{subfigure}
    \hfill
        \begin{subfigure}{0.32\textwidth}
        \centering
        \includegraphics[width=\columnwidth]{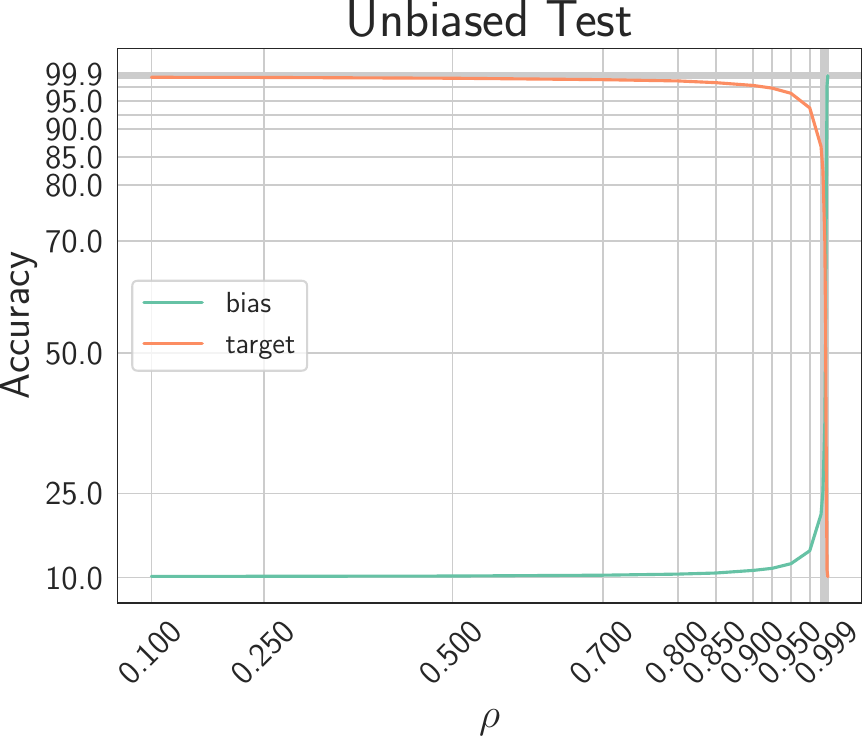}
        \caption{~}
        \label{fig:mnist-unbiased-test}
    \end{subfigure}
    \hfill
    \begin{subfigure}{0.32\textwidth}
        \centering
        \includegraphics[width=\columnwidth]{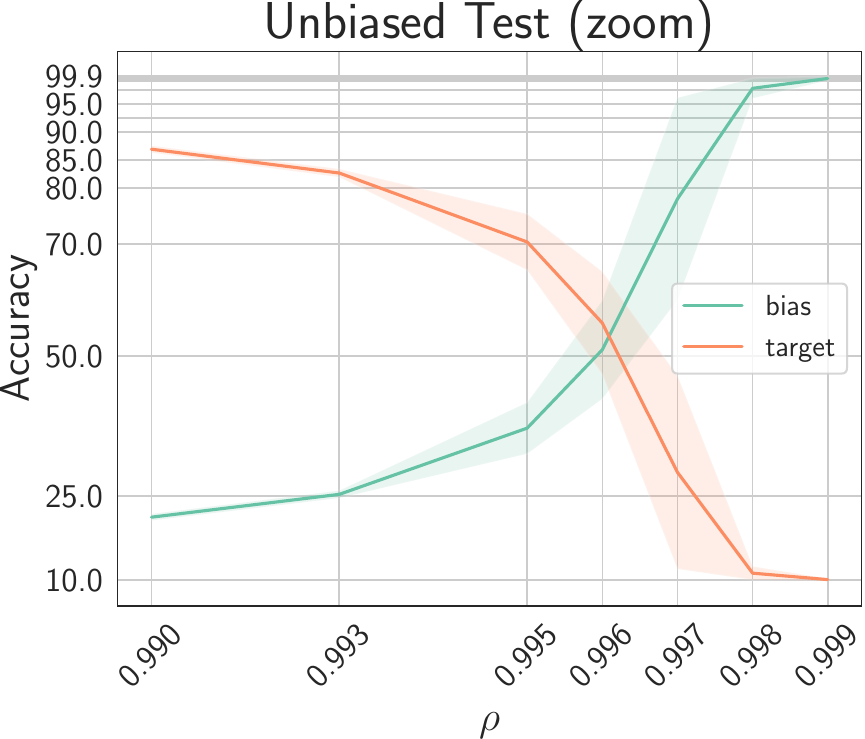}
        \caption{~}
        \label{fig:mnist-unbiased-test-zoom}
    \end{subfigure}
    
    \caption{Biased-MNIST accuracy on the training set (a), and on the unbiased test set (b) and (c). Results are reported in terms of mean and std across three different runs for every value of $\rho$. Given that the number of bias classes (colors) and target classes (digits) is the same, we can compute the bias accuracy by finding the permutation of predicted labels which maximizes the overlap with the ground truth bias labels.}
    \label{fig:mnist-training}
\end{figure*}

Experimental details are provided in the supplementary material. %
The results for Biased-MNIST are presented in Table~\ref{table:mnist-results}. We report the accuracy of the unbiased test set. %
For U-EnD, we evaluate the results employing pseudo-labels computed at different training iterations ($T$) of the biased encoder: at an early stage after 10 epochs, and a late stage at the end of training (80 epochs). In this section, we focus on late-stage pseudo-labeling, while a detailed comparison between the two cases will be the subject of Section~\ref{sec:easy_first}.
Using the unsupervised method we can match the original performance of EnD with the ground-truth bias labels in most settings: this is true especially when the bias is stronger (higher $\rho$ values). This is because in these cases, the bias-capturing models will produce representations strongly biased towards the color, and the pseudo-labels obtained with the bias predictor model will be accurate. On the other hand, a slightly larger gap is observed when there is less correlation between target and bias features. 
This is the most difficult setting for the unsupervised clustering of the bias features: however, a significant improvement to the baseline is always achieved. 
It may be argued that in such cases of weaker bias (or even absence of it), the representations extracted by the biased encoder will be more aligned with the target class rather the the bias features. In this case, the resulting pseudo-labels will be less representative of the actual bias, leading to the disentangling, instead, of the target labels. We identify two worst-case scenarios that might lead to inaccurate pseudo-labels: \emph{i.)} the training set is already unbiased, \emph{ii.)} the pseudo-labels we identify correspond to the target rather than to the bias labels. In this case, applying a debiasing technique might lead to worse performance with respect to the baseline, however, we can avoid this issue thanks to the hyper-parameters optimization policy that we employ. A more detailed analysis of the worst-case settings can be found in the supplementary material. %

\subsection{Easier patterns are preferred}
\label{sec:theoretical-model}

In this section, we analyze how the bias affects the learning process, to provide a better understanding of the bias-capturing model.
By analyzing the training process at different values of $\rho$, we can identify when the color bias shifts from benign to malignant. Fig.~\ref{fig:mnist-training} shows the training accuracy of vanilla models %
trained with different values of $\rho$. Given that, in this case, the number of target classes and the number of different colors (bias classes) are the same, we can compute a bias \emph{pseudo-accuracy}
by finding the permutation of the predicted labels which maximizes the accuracy for the ground truth bias labels: this value gives us an indication of how the final predictions of the model are aligned with the bias. 
From Fig.~\ref{fig:mnist-train} we observe that the target accuracy on the training set is, as expected, close to 100\%, while the bias accuracy is exactly the value of $\rho$, meaning that the models learned to recognize the digit.
This holds also for the unbiased test set (Fig~\ref{fig:mnist-unbiased-test}), where the value $\rho=0.1$. However, if we focus on the higher end of $\rho$ values (most difficult settings) as shown in Fig.~\ref{fig:mnist-unbiased-test-zoom}, we observe a rapid inversion in the trend: the target accuracy decreases, dropping to 10\% for $\rho=0.999$, while the bias accuracy becomes higher, close to 100\% towards the end of the $\rho$ range. In these settings, given the strong correlation between target and bias classes, it is clear that the bias has become easier for the model to learn, and thus malignant. These results can be viewed as further confirmation that neural networks tend to prefer and prioritize the learning of simpler patterns first, as noted by Nam~\emph{et~al.}~\cite{nam2020learning}, \addedrev{Zhang~\emph{et~al.}~ \cite{zhang2023learning}}, and especially
Arpit~\emph{et~al.}~\cite{arpit2017memorization}: it is clear that, to be malignant, a bias must:

\begin{itemize}
    \item be an \emph{easier} pattern to learn that the target features;
    \item have a strong enough correlation with the target task.
\end{itemize}

\subsection{The effect of bias on learning}
\label{sec:bias-theoretical-model}
To provide a more in-depth understanding of the behavior just described and to aid in the explanation of our debiasing technique, we propose a simple yet effective mathematical framework describing the impact of the bias on the learning process. To this end, we want to model the relation among three random variables $T$, $B$, and $Y$, representing the target class, the bias class, and the feature vector extracted by the model, respectively. 

Under the assumption that the target and biases classes are uniformly distributed across the dataset, we have:
\begin{align}
	H(T) &= -\sum_{i=1}^{N_T} P(t_i) \log_2 P(t_i) = \log N_T \nonumber \\
	H(B) &= -\sum_{j=1}^{N_T} P(b_j) \log_2 P(b_j) = \log N_T, \label{eq:entropy_assumption}
\end{align}
where $T$ and $B$ are the random variables associated with the target and the bias class, respectively. Under the assumption of the perfect learner, we must impose
   $H(Y|T) = 0$
(or in other words, the output $y_i$ of the model matches the ground truth value $t_i~\forall i$), and following the steps described in the supplementary material, %
we can write the \emph{normalized} mutual information
\begin{equation}
    \label{eq:MI_perfect}
	\widehat{I}_{perf}(B ; Y) =  \log_{N_T}\left\{N_T\rho\left[\frac{1-\rho}{\rho(N_T-1)}\right]^{1-\rho}\right\}.
\end{equation}
\begin{figure}
    \centering
    \includegraphics[width=0.9\columnwidth]{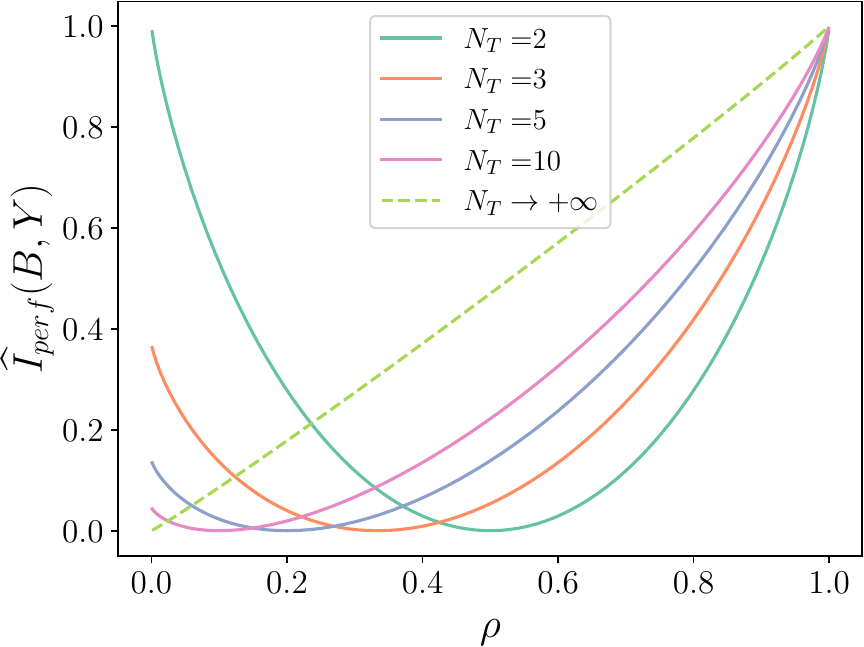}
    \caption{Normalized mutual information in case of perfect learner. As the number of classes $N_T$ increases, the curve smoothens. For low $N_T$ and low $\rho$ values, the anti-correlation phenomenon rises, and the mutual information increases.}
    \label{fig:MIperf_YZ}
\end{figure}
As it is possible to observe in Fig.~\ref{fig:MIperf_YZ}, a clear dependency between $\rho$ and \eqref{eq:MI_perfect} exists. In this case, the biased features and the target ones are in perfect overlap. Nonetheless, in the more general case, the trained model is not a perfect learner, having $H(Y|T) \neq 0$.
The model, in this case, does not correctly classify the target for two reasons.
\begin{enumerate}
	\item It gets confused by the bias features, and it tends to learn to classify samples based on them. We model this tendency of learning biased features with $\phi$, which we call \emph{biasedness} or \emph{unfairness}. The higher the biasedness is, the more the model relies on features that we desire to suppress, inducing bias in the model and for instance error in the model.
	\item Some extra error $\varepsilon$, non-directly related to the bias features, which can be caused, for example, by stochastic unbiased effects, to underfit, or to other high-order dependencies between data. %
\end{enumerate}
We can write the discrete joint probability for $T, B, Y$, composed of the following terms.
\begin{itemize}
    \item When target, bias, and prediction are aligned, the bias is aligned with the target class and correctly classified. Considering that we are not perfect learners, we introduce the error term $\varepsilon$.
    \item When target and bias are aligned but not bias and output, and the prediction is incorrect, it means that the model has not learned the correct feature and the bias is being contrasted.
    \item When target and bias are not aligned, the prediction is incorrect, and bias and output are aligned, it means that the model has learned the bias, introducing the error we target to minimize in this work.
    \item In all the other cases, the error of the model is due to higher-order dependencies, not directly related to the biasedness $\phi$.
\end{itemize}
More formally, we can express the joint probability as:
\begin{align}
	P(T&,B,Y) = \frac{1}{N_T} \cdot \left[\delta_{tby} \rho (1 - \varepsilon) + \bar{\delta}_{ty}\delta_{tb}\bar{\delta}_{by} \frac{(1-\phi)(1-\rho)}{N_T-1} \nonumber \right .\nonumber \\[1em]
	&+ \bar{\delta}_{tb}\bar{\delta}_{ty}\delta_{by}\frac{\phi(1-\rho)}{N_T-1} + \delta_{tb}\bar{\delta}_{ty}\bar{\delta}_{by} \frac{\varepsilon \rho^2}{N_T-2+\rho} \nonumber \\[1em]
	&\left . + \bar{\delta}_{tb}\bar{\delta}_{ty}\bar{\delta}_{by} \frac{\varepsilon \rho (1-\rho)}{(N_T-1)(N_T-2+\rho)} \right],
    \label{eq:Pjointtotal}
\end{align}
where $\delta$ is the Kronecker delta function,\footnote{for easiness of notation we suppress the index $i$: with $\delta_{tby}$ we implicitly intend that bias target and output are aligned to some $i$; hence $\delta_{ti}\delta_{bi}\delta_{yi}$} $\bar{\delta}=1-\delta$ and $\phi,\varepsilon\in[0; 1]$.\footnote{not all the possible combinations are present in the joint probability \eqref{eq:Pjointtotal}: the missing combinations are  considered impossible, like having bias misaligned from the output but target aligned with the bias and with the output of the model (it would correspond to the case $\bar{\delta}_{by}\delta_{ty}\delta_{tb})$} We marginalize \eqref{eq:Pjointtotal} over $T$, obtaining
\begin{align}
	P(B,Y) =& \frac{1}{N_T} \cdot \left\{\delta_{by} [\rho (1 - \varepsilon) + \phi(1-\rho)] + \right .\nonumber \\
 &\left .+\bar{\delta}_{by}\left [\frac{(1-\phi)(1-\rho)}{N_T-1} + \frac{\rho\varepsilon}{N_T-2+\rho}\right ]\right\}
	\label{eq:joint_bz}
\end{align}
from which we compute the normalized mutual information
\begin{align}
    &\widehat{I}(B; Y) = \log_{N_T}\left[\rho (1 - \varepsilon) + \phi(1-\rho)\right]^{\rho (1 - \varepsilon)+ \phi(1-\rho)}\nonumber\\
    &+ \log_{N_T} \left [\frac{(1-\phi)(1-\rho)}{N_T-1} + \frac{\rho\varepsilon}{N_T-2+\rho}\right ]^{(1-\phi)(1-\rho) + \frac{(N_T-1)\rho\varepsilon}{N_T-2+\rho}}\nonumber\\
    &+ 1 + \rho\varepsilon \left(\frac{N_T-1}{N_T-2-\rho} -1 \right).\label{eq:MIcool}
\end{align}
\begin{figure}
    \centering
    \includegraphics[width=0.97\columnwidth,trim={0 0 0 30},clip]{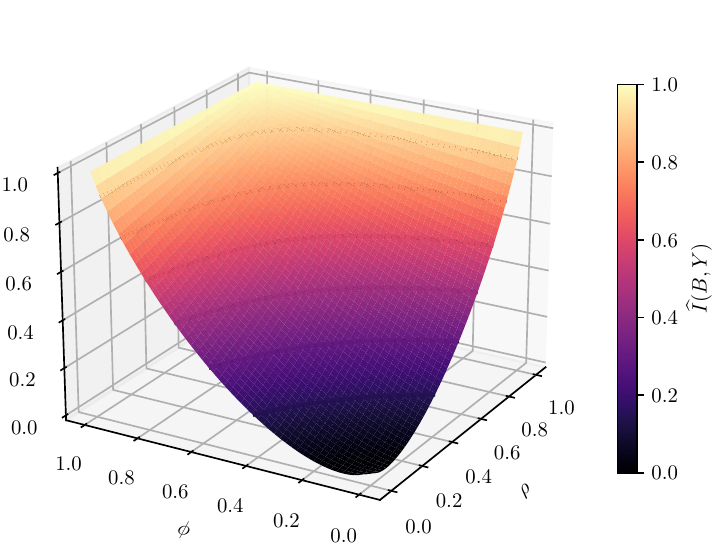}
    \caption{Normalized mutual information between B and Y, in the case $\varepsilon=0$ and $N_T=10$. In light black, on top of the graph, level curves are drawn. In high $\rho$ regions, the mutual information is high regardless of the bias tendency $\phi$ value, motivating the difficulty in learning the disentangling from biased features. A projection to a lower $\rho$ makes the problem of minimizing $\phi$ can drive the disentangling.}
    \label{fig:MI_YZ}
\end{figure}
The complete derivation can be found in the supplementary material %
and the plot for \eqref{eq:MIcool} in the case $\varepsilon=0$ is displayed in Figure~\ref{fig:MI_YZ}. 
Interestingly, we observe that for high values of $\rho$, the mutual information between the features
learned by the model and the bias is high, independently from the biasedness $\phi$.
Typical learning scenarios, indeed, work in this region, where it is extremely challenging to optimize over $\phi$. On the contrary, with a lower $\rho$, the effect of $\phi$ appears evident. Towards this end, relying on a (relatively small) balanced validation set (hence, with low $\rho$) is extremely important to enhance the lowering of $\phi$. Directly optimizing over $\phi$ is in general not a doable strategy as the mutual information between B and Y in the typical training scenarios is very high; hence while bias removal can succeed, it will be extremely difficult to disentangle the biased features from the unbiased ones, without harming the performance.

\subsection{Easier patterns are learned first}
\label{sec:easy_first}
Besides being easier to learn than the target task, we also find that biases tend to be learned in the first epochs. This is also evident when looking at the results in Table~\ref{table:mnist-results}, with $T=10$: using an early bias predictor results in more precise pseudo-labels, especially when $\rho$ is lower. In order to measure how biased the encoder is, we employ the theoretical model we presented in Eq.~\eqref{eq:joint_bz} to compute $\phi$ (the derivation can be found in the supplementary material). %
In Fig.~\ref{fig:bias-tendency} we show the mean value of $\phi$ measured at different training iterations (we assume $\varepsilon=0$). As expected, the models tend to show a stronger tendency towards bias when $\rho$ is higher. Interestingly, looking at the dynamics it is also clear that this behavior is exhibited predominantly in earlier epochs.
Under certain conditions, i.e. when the correlation between target and bias is not as strong, the optimization process can escape the local minimum corresponding to a biased model. These findings are also confirmed by the related literature~\cite{nam2020learning, arpit2017memorization}. 

\begin{figure*}
    \centering
    \includegraphics[width=\textwidth]{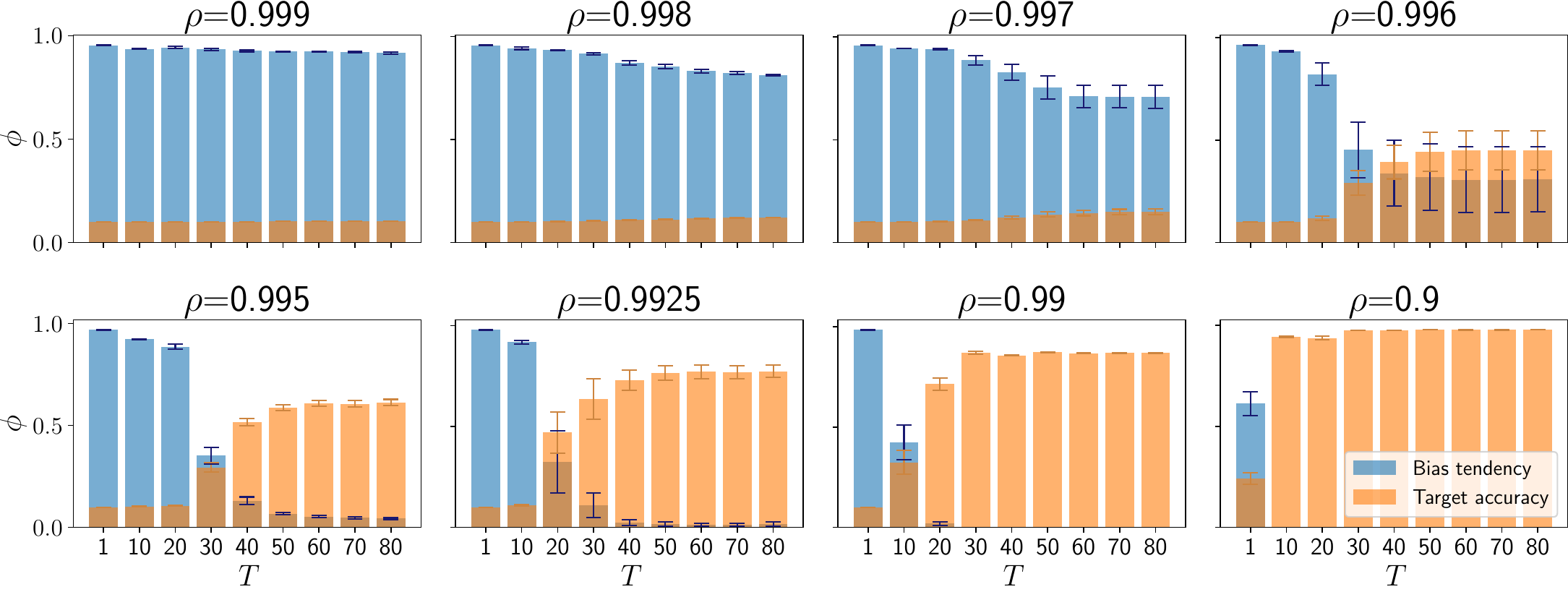}
    \caption{Biasedness/Unfairness ($\phi$), or tendency towards learning bias features, in terms of mean and std computed across three independent runs for different values of $\rho$.}
    \label{fig:bias-tendency}
\end{figure*}

\section{Real world experiments}
\label{sec:experiments}

\begin{table}
    \centering
    \begin{tabular}{l@{\hspace{1\tabcolsep}} l@{\hspace{1\tabcolsep}} c c c}
        \toprule
        & Method & Unbiased & Bias-conflicting\\
        \midrule
        \multirow{6}{*}{\rotatebox{90}{Hair Color}}
        & Vanilla~\cite{nam2020learning} & 70.25\std{0.35} & 52.52\std{0.19} \\
        & Group DRO~\cite{sagawa2019distributionally} & \underline{85.43}\std{0.53} & \underline{83.40} \std{0.67}\\
        & EnD~\cite{tartaglione2021end} & \textbf{91.21}\std{0.22} & \textbf{87.45}\std{1.06}\\\addlinespace[1ex]
        \cline{2-4}\addlinespace[1ex]
        & LfF$^\dagger$~\cite{nam2020learning} & \underline{84.24}\std{0.37} & \textbf{81.24}\std{1.38} \\
        & U-EnD$^\dagger$ ($T$=50) & 83.97\std{2.90} & \underline{74.18}\std{6.07} \\
        & U-EnD$^\dagger$ ($T$=30) & \textbf{84.39}\std{2.38} & 72.53\std{4.47} \\
        \midrule
        \midrule
        \multirow{7}{*}{\rotatebox{90}{\small Heavy Makeup}}
        & Vanilla~\cite{nam2020learning} & 62.00\std{0.02} & 33.75\std{0.28} \\
        & Group DRO~\cite{sagawa2019distributionally} & \underline{64.88}\std{0.42} & \underline{50.24}\std{0.68} \\
        & EnD~\cite{tartaglione2021end} & \textbf{75.93}\std{1.31} & \textbf{53.70}\std{5.24}  \\\addlinespace[1ex]
        \cline{2-4}\addlinespace[1ex]
        & LfF$^\dagger$~\cite{nam2020learning} & 66.20\std{1.21} & \textbf{45.48}\std{4.33} \\
        & U-EnD$^\dagger$ ($T$=50) & \textbf{72.22}\std{0.00} & \underline{44.44}\std{0.00} \\
        & U-EnD$^\dagger$ ($T$=30) & \underline{67.59}\std{3.46} & 35.19\std{6.93} \\
        \bottomrule
    \end{tabular}
    \caption{\textbf{Performance on CelebA.} Techniques that can be used in an unsupervised way are denoted with $^\dagger$.}
    \label{table:celeba-results}
\end{table}

\begin{table}
    \centering
    \resizebox{0.9\columnwidth}{!}{%
    \begin{tabular}{@{}l@{\hspace{0.8\tabcolsep}} l c c c c}
        \toprule
        &\multirow{2}{*}{Method}& \multicolumn{2}{c}{Trained on EB1} & \multicolumn{2}{c}{Trained on EB2} \\
        & & EB2 & Test & EB1 & Test \\
        \midrule
        \multirow{5}{*}{\rotatebox{90}{Learn Gender}}
        & Vanilla~\cite{Kim_2019_CVPR}                                & 59.86             & 84.42             & 57.84             & 69.75 \\
        & BlindEye~\cite{alvi2018turning}         & 63.74             & 85.56             & 57.33             & 69.90 \\
        & LNL~\cite{Kim_2019_CVPR}  & \underline{68.00} & 86.66 & 64.18 & 74.50 \\
        & EnD~\cite{tartaglione2021end}   & 65.49\std{0.81} & \underline{87.15}\std{0.31}  & \underline{69.40}\std{2.01}   & \underline{78.19}\std{1.18} \\
        & U-EnD$^\dagger$ ($T$=50) & \textbf{81.32}\std{2.17} & \textbf{90.98}\std{0.46} & \textbf{78.10}\std{0.70} & \textbf{83.03}\std{0.45} \\
        \midrule
        \multirow{5}{*}{\rotatebox{90}{Learn Age}} 
        &Vanilla~\cite{Kim_2019_CVPR}                                & 54.30 & 77.17 & 48.91 & 61.97 \\
        &BlindEye~\cite{alvi2018turning}        & 66.80 & 75.13 & {64.16} & 62.40 \\
        &LNL~\cite{Kim_2019_CVPR} & 65.27 & 77.43 & 62.18 & {63.04} \\
        & EnD~\cite{tartaglione2021end} & \underline{76.04}\std{0.25} & \underline{80.15}\std{0.96}  & \textbf{74.25}\std{2.26}  & \textbf{78.80}\std{1.48} \\  
        & U-EnD$^\dagger$ ($T$=50) & \textbf{80.41}\std{2.96} & \textbf{83.43}\std{2.49} & \underline{70.82}\std{1.04} & \underline{76.09}\std{0.91} \\
        \bottomrule
    \end{tabular}
    }
    \caption{\textbf{Performance on IMDB Face.} When gender is learned, age is the bias, and when age is learned the gender is the bias. Techniques that can be used in an unsupervised way are denoted with $^\dagger$.}
    \label{table:imdb-results}
\end{table}

In this section, we present the experiments we performed on real-world datasets, where biases can be much harder to identify and overcome. In these experiments, we deal with different common kinds of biases in real datasets, 
specifically gender bias and age bias in facial images.
We use two very common datasets for facial recognition tasks, CelebA~\cite{liu2015faceattributes} and the IMDB face dataset~\cite{Rothe-IJCV-2018}. 

\subsection{Setup}
We use the common convolutional architecture ResNet-18 proposed by \cite{he2016deep}. 
For CelebA and IMDB, the network is pre-trained on ImageNet, except for the last fully connected layer. The same architecture is used both for training the biased encoder and the unbiased model. As in the previous experiments, the EnD regularization is applied on the encoder embeddings (average pooling layer). More experimental details are provided in the supplementary material.\footnote{Source code will be made publicly available upon publication}

\subsection{CelebA}
\begin{figure}
    \centering
    \includegraphics[width=\columnwidth]{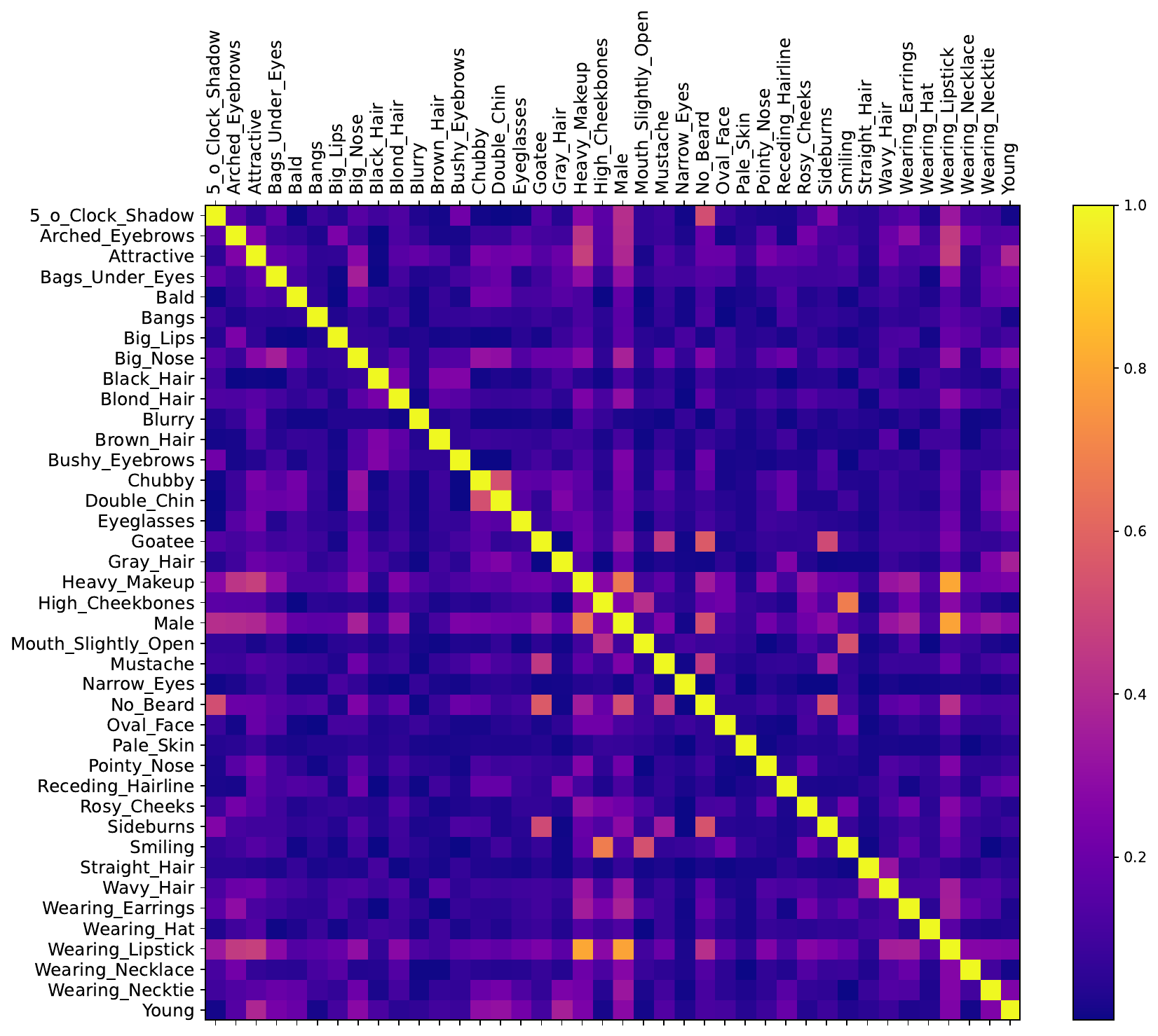}
    \caption{\added{Correlation (absolute value) between the different attributes in the CelebA dataset. It can be clearly observed how facial images exhibit multiple biases that might be better captured by unsupervised techniques.}}
    \label{fig:celeba-correlation}
\end{figure}

CelebA~\cite{liu2015faceattributes} is a widely known facial image dataset, comprising 202,599 images. It is built for face recognition tasks and provides 40 binary attributes for every image. 
Similarly to \cite{nam2020learning}, we use attributes indicating the hair color and the presence of makeup as target attributes while using gender as bias. The reason is that there is a high correlation between these choices of attributes, with most of the women in the dataset being blond or having heavy facial makeup. \added{However, it is worth noting that these are not the only correlated attributes in the dataset, as Fig.~\ref{fig:celeba-correlation} shows.} We utilize the official train-validation split for training (162,770 images) and testing (19,867 images). As in~\cite{nam2020learning} we build two different testing datasets starting from the original one: an \emph{unbiased} set, in which we select the same number of samples for every possible pair of target and bias attributes, and a \emph{bias-conflicting}, where we only select samples where the values of target and bias attributes are different (e.g. men with blonde hair, women without makeup, and so on). This allows us to evaluate the model performance in the most difficult setting where the target is not aligned with the bias. Experimental details are provided in
the supplementary material. %

\textbf{Results.} 
Denoting with $t$ and $b$ the target and bias attribute respectively, we computed the final accuracy as average accuracy across all the $(t, b)$ pairs, as in~\cite{nam2020learning}. We report the results in Table~\ref{table:celeba-results}. Results are reported for both the target attributes hair color and makeup. Techniques that can be used in an unsupervised manner are denoted with $^\dagger$. We report baseline results (vanilla) and we observe how vanilla models suffer significantly from the presence of the bias, scoring a quite low accuracy (especially since this is a binary task). This is evident in the bias-conflicting set, where the performance is close random-guess on hair color prediction, and even lower on the makeup detection. We report reference results~\cite{nam2020learning} of other debiasing algorithms, specifically Group~DRO~\cite{sagawa2019distributionally}, LfF~\cite{nam2020learning} and EnD~\cite{tartaglione2021end}. Focusing on supervised techniques (Group DRO and EnD) we observe a significant increase in performance, in both the tasks and test set combinations. For the unsupervised methods, we report the results of our U-EnD at different $T$ of the biased encoder, as done in Table~\ref{table:mnist-results}, and compare them to LfF. We achieve better performance than the vanilla baseline in all settings, even though we still observe a gap with respect to the fully supervised techniques. The same observation can be made for LfF, which in general performs better on the harder cases in the bias-conflicting set, while U-EnD provides better performance in the more general case of the unbiased test set. 
The observed results are similar to the lower $\rho$ settings of BiasedMNIST: the amount of biased information is sufficient for it to be considered as a malignant bias, although it becomes slightly harder to perform pseudo-labeling in the biased encoder latent space. However, the assumptions we make in Sec.~\ref{sec:phase3} about the pseudo-labeling accuracy hold, resulting in better results with respect to the baseline models.

\subsection{IMDB Face}
The IMDB Face dataset~\cite{Rothe-IJCV-2018} contains 460,723 face images annotated with age and gender information. This dataset is known to present noisy labels for age, thus \cite{Kim_2019_CVPR} perform a cleaning step to filter out noisy labels from the data, by using a model trained on the Audience benchmark dataset~\cite{eidinger2014age}. Only the samples for which the original age value matches the predicted one are considered. As done in~\cite{Kim_2019_CVPR}, the filtered dataset is then split with an 80\%-20\% ratio between training and testing data. To construct a biased training set, the full training set subsequently split into two \emph{extreme bias} (EB) sets: \emph{EB1}, which contains women only in the age range 0-29 and men with age 40+, and \emph{EB2} which contains man only in the age range 0-29 and women with age 40+. This way, a strong correlation between age and gender is obtained: when training age prediction, the bias is given by the gender, and when recognizing the gender, the bias is given by the age. 
Experimental details are provided in the supplementary material. %

\textbf{Results.}
We report the results on the IMDB Face dataset in Table~\ref{table:imdb-results}, with regards to both gender and age prediction. Besides the test set, every model is also tested on the opposite EB set, to better evaluate the debiasing performance. As in the previous experiments, we use $^\dagger$ to denote the techniques that can be used in an unsupervised way. Focusing on the supervised techniques, we observe a significant improvement with respect to the baselines, especially with EnD and LNL, across the different combinations of test sets and tasks. Interestingly, in this case, we can achieve even better results when employing the U-EnD approach, contrary to the CelebA results. Especially for learning gender, we notice the performance is noticeably higher than the best-supervised results. This might be due to the noisy age labels in the dataset, and even if the described cleaning procedure is applied some labels could still be incorrect. With pseudo-labeling, on the other hand, we do not make use of the provided labels. This might be confirmed by the performances obtained when training for age prediction. As the gender label is of course far less noisy than the age, the performance gap between EnD and U-EnD is far less noticeable. \added{Another potential explanation is that, as shown in Fig.~\ref{fig:celeba-correlation} for CelebA, multiple biases could be present in the images also for IMDB (even if they are not annotated in the dataset). Thus, an unsupervised method may be better suited for handling this case.} We believe these results are very important, as they show that it is sometimes possible to achieve better results with unsupervised approaches. 

\subsection{\added{Corrupted CIFAR-10}}
\added{Corrupted CIFAR-10~\cite{Hendrycks2019} is built from the CIFAR-10 dataset, by applying a textural bias (e.g. brightness, frost corruption, etc.) to the different classes. Similarly to Biased-MNIST, we test different levels of corruption. For this dataset, an unbiased validation set is not available, and so for training the bias-capturing model, we employ the generalized cross-entropy loss (GCE)~\cite{nam2020learning}, which favors the selection of biased features. Similarly to Biased-MNIST, we pick four different levels of difficulty, given by the ratio of bias-aligned and bias-conflicting samples, comprised in $~{\{0.5\%, 1\%, 2\%, 5\%\}}$, as in~\cite{nam2020learning}.}

\textbf{Results.} \added{The results are reported in Tab.~\ref{table:cifar10c-results}. We compare with different baselines~\cite{wang2018learning, bahng2019rebias, nam2020learning, lee2021learning} and report the results of fully-supervised and unsupervised EnD. Also in this case, with U-EnD, we can match (and slightly surpass with a ratio of one) the performance of the supervised baseline EnD. For the ratio of five, we can observe that the results achieved by U-EnD are slightly lower than EnD. This can be due to the lack of a proper validation set, as the bias is more subtle. However, we still observe a significant increase with respect to the vanilla baseline.}

\begin{table}
    \centering
    \resizebox{0.9\columnwidth}{!}{%
    \begin{tabular}{l c c c c}
        \toprule
        \multirow{2}{*}{Method} & \multicolumn{4}{c}{Ratio (\%)}\\
                                & 0.5 & 1.0 & 2.0 & 5.0 \\ 
        \toprule
        Vanilla~\cite{lee2021learning} & 23.08\std{1.25} & 25.82\std{0.33} & 30.06\std{0.71} & 39.42\std{0.64} \\
        EnD~\cite{tartaglione2021end} & \textbf{28.57}\std{0.80} & \textbf{29.92}\std{0.47} & \textbf{34.73}\std{1.17} & \textbf{44.73}\std{0.20} \\
        \midrule
        HEX$^\dagger$~\cite{wang2018learning} & 13.87\std{0.06} & 14.81\std{0.42} & 15.20\std{0.54} & 16.04\std{0.63} \\
        ReBias$^\dagger$~\cite{bahng2019rebias} &22.27\std{0.41} & 25.72\std{0.20} & 31.66\std{0.43} & \underline{43.43}\std{0.41} \\
        LfF$^\dagger$~\cite{nam2020learning} & \textbf{28.57}\std{1.30} & \textbf{33.07}\std{0.77} & \textbf{39.91}\std{0.30} & \textbf{50.27}\std{1.56} \\
        U-EnD$^\dagger$ ($T$=500) & \underline{27.49}\std{0.96} & \underline{30.33}\std{0.16} & \underline{32.39}\std{1.80} & 41.26\std{0.81} \\
        \bottomrule
    \end{tabular}
    }
    \caption{\added{\textbf{Corrupted CIFAR-10 accuracy} with different corruption ratio (\%). Techniques that can be used in an unsupervised way are denoted with $^\dagger$.}}
    \label{table:cifar10c-results}
\end{table}

\section{Conclusion}
\label{sec:conclusion}

In this work, we proposed an unsupervised framework for learning unbiased representations from biased data. Our framework consists of three separate steps: \emph{i.)} training a bias-capturing model \emph{ii.)} training a bias predictor \emph{iii.)} training an unbiased model. 
We obtain the bias-capturing model by exploiting the tendency of optimization and neural networks to prefer simpler patterns over more complex ones. We show that when such patterns exist, and they represent spurious correlations with the target features, then the model will rely on these confounding factors. 
In Section~\ref{sec:phase1} we show that such cases correspond to converging towards a local minimum for the target tasks, which however provides an optimal solver if we consider instead the task of predicting bias features. Furthermore, we the theoretical framework presented in Section~\ref{sec:bias-theoretical-model} we are able to empirically measure the biasness of the model, which can be a useful insight in dealing with potentially biased data. We leverage these findings for adapting fully supervised debiasing techniques in an unsupervised context, via a pseudo-labeling step on the biased latent space. Thanks to this approach, we are able to use state-of-the-art debiasing algorithms such as EnD, which is a strong regularizer for driving the model towards the selection of unbiased features. With experiments on real-world data, we also show how sometimes it is even possible to achieve better results with an unsupervised approach, addressing the issue of noisy labels (both regarding target classes and bias classes) in datasets, \added{and taking into account instances in which multiple biases exist}.
We believe that our approach could be of potentially great interest for other researchers in the area, and also for practical applications, as it can be easily adapted to different techniques for building bias-capturing models and for obtaining unbiased predictors.

\section*{Acknowledgements}
This paper has been supported by the French National Research Agency (ANR) in the framework of
the JCJC project ``BANERA'' ANR-24-CE23-4369, and by Hi!PARIS Center on Data Analytics and Artificial Intelligence. This work was granted access to the HPC/AI resources of IDRIS under the allocation 2023-AD011013473R1 made by GENCI.

\bibliographystyle{IEEEtran}
\bibliography{main}

\begin{thebibliography}{10}
\providecommand{\url}[1]{#1}
\csname url@samestyle\endcsname
\providecommand{\newblock}{\relax}
\providecommand{\bibinfo}[2]{#2}
\providecommand{\BIBentrySTDinterwordspacing}{\spaceskip=0pt\relax}
\providecommand{\BIBentryALTinterwordstretchfactor}{4}
\providecommand{\BIBentryALTinterwordspacing}{\spaceskip=\fontdimen2\font plus
\BIBentryALTinterwordstretchfactor\fontdimen3\font minus \fontdimen4\font\relax}
\providecommand{\BIBforeignlanguage}[2]{{%
\expandafter\ifx\csname l@#1\endcsname\relax
\typeout{** WARNING: IEEEtran.bst: No hyphenation pattern has been}%
\typeout{** loaded for the language `#1'. Using the pattern for}%
\typeout{** the default language instead.}%
\else
\language=\csname l@#1\endcsname
\fi
#2}}
\providecommand{\BIBdecl}{\relax}
\BIBdecl

\bibitem{voulodimos2018deep}
A.~Voulodimos, N.~Doulamis, A.~Doulamis, and E.~Protopapadakis, ``Deep learning for computer vision: A brief review,'' \emph{Computational intelligence and neuroscience}, vol. 2018, 2018.

\bibitem{NEURIPS2020_1457c0d6}
T.~Brown, B.~Mann, N.~Ryder, M.~Subbiah, J.~D. Kaplan, P.~Dhariwal, A.~Neelakantan, P.~Shyam, G.~Sastry, A.~Askell, S.~Agarwal, A.~Herbert-Voss, G.~Krueger, T.~Henighan, R.~Child, A.~Ramesh, D.~Ziegler, J.~Wu, C.~Winter, C.~Hesse, M.~Chen, E.~Sigler, M.~Litwin, S.~Gray, B.~Chess, J.~Clark, C.~Berner, S.~McCandlish, A.~Radford, I.~Sutskever, and D.~Amodei, ``Language models are few-shot learners,'' in \emph{Advances in Neural Information Processing Systems}, H.~Larochelle, M.~Ranzato, R.~Hadsell, M.~F. Balcan, and H.~Lin, Eds., vol.~33.\hskip 1em plus 0.5em minus 0.4em\relax Curran Associates, Inc., 2020, pp. 1877--1901.

\bibitem{dosovitskiy2020vit}
A.~Dosovitskiy, L.~Beyer, A.~Kolesnikov, D.~Weissenborn, X.~Zhai, T.~Unterthiner, M.~Dehghani, M.~Minderer, G.~Heigold, S.~Gelly, J.~Uszkoreit, and N.~Houlsby, ``An image is worth 16x16 words: Transformers for image recognition at scale,'' \emph{ICLR}, 2021.

\bibitem{laumer2015impact}
S.~Laumer, C.~Maier, and A.~Eckhardt, ``The impact of business process management and applicant tracking systems on recruiting process performance: an empirical study,'' \emph{Journal of Business Economics}, vol.~85, no.~4, pp. 421--453, 2015.

\bibitem{hleg2019ethics}
\BIBentryALTinterwordspacing
E.~C.~A. HLEG), \emph{Ethics guidelines for trustworthy AI}.\hskip 1em plus 0.5em minus 0.4em\relax High-Level Expert Group on Artificial Intelligence, 2019. [Online]. Available: \url{https://ec.europa.eu/digital-single-market/en/news/ethics-guidelines-trustworthy-ai}
\BIBentrySTDinterwordspacing

\bibitem{zhang2019artificial}
B.~Zhang and A.~Dafoe, ``Artificial intelligence: American attitudes and trends,'' \emph{Available at SSRN 3312874}, 2019.

\bibitem{schramowski2020making}
P.~{Schramowski}, W.~{Stammer}, S.~{Teso}, A.~{Brugger}, F.~{Herbert}, X.~{Shao}, H.-G. {Luigs}, A.-K. {Mahlein}, and K.~{Kersting}, ``Making deep neural networks right for the right scientific reasons by interacting with their explanations,'' \emph{Nature Machine Intelligence}, vol.~2, no.~8, pp. 476--486, 2020.

\bibitem{stock2020eccv}
\BIBentryALTinterwordspacing
P.~Stock and M.~Ciss{\'{e}}, ``Convnets and imagenet beyond accuracy: Understanding mistakes and uncovering biases,'' in \emph{Computer Vision - {ECCV} 2018 - 15th European Conference, Munich, Germany, September 8-14, 2018, Proceedings, Part {VI}}, ser. Lecture Notes in Computer Science, V.~Ferrari, M.~Hebert, C.~Sminchisescu, and Y.~Weiss, Eds., vol. 11210.\hskip 1em plus 0.5em minus 0.4em\relax Springer, 2018, pp. 504--519. [Online]. Available: \url{https://doi.org/10.1007/978-3-030-01231-1\_31}
\BIBentrySTDinterwordspacing

\bibitem{teso2019aies_XIML}
S.~Teso and K.~Kersting, ``Explanatory interactive machine learning,'' in \emph{Proceedings of the 2nd AAAI/ACM Conference on AI, Ethics, and Society (AIES)}, 2019.

\bibitem{revisetool}
A.~Wang, A.~Narayanan, and O.~Russakovsky, ``{REVISE}: A tool for measuring and mitigating bias in visual datasets,'' \emph{European Conference on Computer Vision (ECCV)}, 2020.

\bibitem{barbano2021bridging}
C.~A. Barbano, E.~Tartaglione, and M.~Grangetto, ``Bridging the gap between debiasing and privacy for deep learning,'' in \emph{Proceedings of the IEEE/CVF International Conference on Computer Vision (ICCV) Workshops}, October 2021, pp. 3806--3815.

\bibitem{nam2020learning}
J.~Nam, H.~Cha, S.~Ahn, J.~Lee, and J.~Shin, ``Learning from failure: Training debiased classifier from biased classifier,'' in \emph{Advances in Neural Information Processing Systems}, 2020.

\bibitem{Kim_2019_CVPR}
B.~Kim, H.~Kim, K.~Kim, S.~Kim, and J.~Kim, ``Learning not to learn: Training deep neural networks with biased data,'' in \emph{The IEEE Conference on Computer Vision and Pattern Recognition (CVPR)}, June 2019.

\bibitem{zhao2021learning}
\BIBentryALTinterwordspacing
B.~Zhao, C.~Chen, Q.~Ju, and S.~Xia, ``Learning debiased models with dynamic gradient alignment and bias-conflicting sample mining,'' 11 2021. [Online]. Available: \url{https://arxiv.org/abs/2111.13108v1}
\BIBentrySTDinterwordspacing

\bibitem{lee2021learning}
\BIBentryALTinterwordspacing
J.~Lee, E.~Kim, J.~Lee, J.~Lee, and J.~Choo, ``Learning debiased representation via disentangled feature augmentation,'' 2021. [Online]. Available: \url{https://openreview.net/forum?id=-oUhJJILWHb https://github.com/kakaoenterprise/Learning-Debiased-Disentangled}
\BIBentrySTDinterwordspacing

\bibitem{tartaglione2021end}
E.~Tartaglione, C.~A. Barbano, and M.~Grangetto, ``End: Entangling and disentangling deep representations for bias correction,'' in \emph{Proceedings of the IEEE/CVF Conference on Computer Vision and Pattern Recognition (CVPR)}, June 2021, pp. 13\,508--13\,517.

\bibitem{dufumier2021brain}
\BIBentryALTinterwordspacing
B.~Dufumier, P.~Gori, I.~Battaglia, J.~Victor, A.~Grigis, and E.~Duchesnay, ``Benchmarking cnn on 3d anatomical brain mri: Architectures, data augmentation and deep ensemble learning,'' \emph{NeuroImage}, 6 2021. [Online]. Available: \url{https://arxiv.org/abs/2106.01132v1}
\BIBentrySTDinterwordspacing

\bibitem{alvi2018turning}
M.~Alvi, A.~Zisserman, and C.~Nell{\aa}ker, ``Turning a blind eye: Explicit removal of biases and variation from deep neural network embeddings,'' in \emph{Proceedings of the European Conference on Computer Vision (ECCV)}, 2018, pp. 0--0.

\bibitem{bahng2019rebias}
H.~Bahng, S.~Chun, S.~Yun, J.~Choo, and S.~J. Oh, ``Learning de-biased representations with biased representations,'' in \emph{International Conference on Machine Learning (ICML)}, 2020.

\bibitem{cadene2019rubi}
R.~Cadene, C.~Dancette, M.~Cord, D.~Parikh \emph{et~al.}, ``Rubi: Reducing unimodal biases for visual question answering,'' in \emph{Advances in neural information processing systems}, 2019, pp. 841--852.

\bibitem{arpit2017memorization}
D.~Arpit, S.~Jastrzundefinedbski, N.~Ballas, D.~Krueger, E.~Bengio, M.~S. Kanwal, T.~Maharaj, A.~Fischer, A.~Courville, Y.~Bengio, and S.~Lacoste-Julien, ``A closer look at memorization in deep networks,'' in \emph{Proceedings of the 34th International Conference on Machine Learning - Volume 70}, ser. ICML'17.\hskip 1em plus 0.5em minus 0.4em\relax JMLR.org, 2017, p. 233–242.

\bibitem{torralba2011unbiased}
A.~Torralba, A.~A. Efros \emph{et~al.}, ``Unbiased look at dataset bias.'' in \emph{CVPR}.\hskip 1em plus 0.5em minus 0.4em\relax Citeseer, 2011, p.~7.

\bibitem{Khosla2012UndoingTD}
A.~Khosla, T.~Zhou, T.~Malisiewicz, A.~A. Efros, and A.~Torralba, ``Undoing the damage of dataset bias,'' in \emph{ECCV}, 2012.

\bibitem{tommasi2017deeper}
T.~Tommasi, N.~Patricia, B.~Caputo, and T.~Tuytelaars, ``A deeper look at dataset bias,'' in \emph{Domain adaptation in computer vision applications}.\hskip 1em plus 0.5em minus 0.4em\relax Springer, 2017, pp. 37--55.

\bibitem{gupta2018robot}
A.~Gupta, A.~Murali, D.~P. Gandhi, and L.~Pinto, ``Robot learning in homes: Improving generalization and reducing dataset bias,'' in \emph{Advances in Neural Information Processing Systems}, 2018, pp. 9094--9104.

\bibitem{song2017machine}
C.~Song, T.~Ristenpart, and V.~Shmatikov, ``Machine learning models that remember too much,'' in \emph{Proceedings of the 2017 ACM SIGSAC Conference on Computer and Communications Security}.\hskip 1em plus 0.5em minus 0.4em\relax ACM, 2017, pp. 587--601.

\bibitem{beutel2019putting}
A.~Beutel, J.~Chen, T.~Doshi, H.~Qian, A.~Woodruff, C.~Luu, P.~Kreitmann, J.~Bischof, and E.~H. Chi, ``Putting fairness principles into practice: Challenges, metrics, and improvements,'' in \emph{Proceedings of the 2019 AAAI/ACM Conference on AI, Ethics, and Society}, 2019, pp. 453--459.

\bibitem{xu2018fairgan}
D.~Xu, S.~Yuan, L.~Zhang, and X.~Wu, ``Fairgan: Fairness-aware generative adversarial networks,'' in \emph{2018 IEEE International Conference on Big Data (Big Data)}.\hskip 1em plus 0.5em minus 0.4em\relax IEEE, 2018, pp. 570--575.

\bibitem{sattigeri2018fairness}
P.~Sattigeri, S.~C. Hoffman, V.~Chenthamarakshan, and K.~R. Varshney, ``Fairness gan,'' \emph{arXiv preprint arXiv:1805.09910}, 2018.

\bibitem{madras2018learning}
D.~Madras, E.~Creager, T.~Pitassi, and R.~Zemel, ``Learning adversarially fair and transferable representations,'' \emph{arXiv preprint arXiv:1802.06309}, 2018.

\bibitem{Xie2017ControllableIT}
Q.~Xie, Z.~Dai, Y.~Du, E.~Hovy, and G.~Neubig, ``Controllable invariance through adversarial feature learning,'' in \emph{NIPS}, 2017.

\bibitem{wang2019iccv}
T.~Wang, J.~Zhao, M.~Yatskar, K.-W. Chang, and V.~Ordonez, ``Balanced datasets are not enough: Estimating and mitigating gender bias in deep image representations,'' in \emph{International Conference on Computer Vision (ICCV)}, October 2019.

\bibitem{ClarkYZ19}
\BIBentryALTinterwordspacing
C.~Clark, M.~Yatskar, and L.~Zettlemoyer, ``Don't take the easy way out: Ensemble based methods for avoiding known dataset biases,'' in \emph{Proceedings of the 2019 Conference on Empirical Methods in Natural Language Processing and the 9th International Joint Conference on Natural Language Processing, {EMNLP-IJCNLP} 2019, Hong Kong, China, November 3-7, 2019}, K.~Inui, J.~Jiang, V.~Ng, and X.~Wan, Eds.\hskip 1em plus 0.5em minus 0.4em\relax Association for Computational Linguistics, 2019, pp. 4067--4080. [Online]. Available: \url{https://doi.org/10.18653/v1/D19-1418}
\BIBentrySTDinterwordspacing

\bibitem{wang2020fair}
Z.~Wang, K.~Qinami, I.~Karakozis, K.~Genova, P.~Nair, K.~Hata, and O.~Russakovsky, ``Towards fairness in visual recognition: Effective strategies for bias mitigation,'' in \emph{IEEE/CVF Conference on Computer Vision and Pattern Recognition (CVPR)}, 2020.

\bibitem{sagawa2019distributionally}
S.~Sagawa, P.~W. Koh, T.~B. Hashimoto, and P.~Liang, ``Distributionally robust neural networks,'' in \emph{International Conference on Learning Representations}, 2019.

\bibitem{barbano2023unbiased}
\BIBentryALTinterwordspacing
C.~A. Barbano, B.~Dufumier, E.~Tartaglione, M.~Grangetto, and P.~Gori, ``Unbiased supervised contrastive learning,'' in \emph{The Eleventh International Conference on Learning Representations}, 2023. [Online]. Available: \url{https://openreview.net/forum?id=Ph5cJSfD2XN}
\BIBentrySTDinterwordspacing

\bibitem{wang2018learning}
\BIBentryALTinterwordspacing
H.~Wang, Z.~He, Z.~L. Lipton, and E.~P. Xing, ``Learning robust representations by projecting superficial statistics out,'' in \emph{International Conference on Learning Representations}, 2019. [Online]. Available: \url{https://openreview.net/forum?id=rJEjjoR9K7}
\BIBentrySTDinterwordspacing

\bibitem{haralick1973textural}
R.~M. Haralick, K.~Shanmugam, and I.~Dinstein, ``Textural features for image classification,'' \emph{IEEE Transactions on Systems, Man, and Cybernetics}, vol. SMC-3, no.~6, pp. 610--621, 1973.

\bibitem{lam1996glcm}
S.-C. Lam, ``Texture feature extraction using gray level gradient based co-occurence matrices,'' in \emph{1996 IEEE International Conference on Systems, Man and Cybernetics. Information Intelligence and Systems (Cat. No.96CH35929)}, vol.~1, 1996, pp. 267--271 vol.1.

\bibitem{hendricks2018women}
L.~A. Hendricks, K.~Burns, K.~Saenko, T.~Darrell, and A.~Rohrbach, ``Women also snowboard: Overcoming bias in captioning models,'' in \emph{European Conference on Computer Vision}.\hskip 1em plus 0.5em minus 0.4em\relax Springer, 2018, pp. 793--811.

\bibitem{Ross2017RightFT}
A.~Ross, M.~Hughes, and F.~Doshi-Velez, ``Right for the right reasons: Training differentiable models by constraining their explanations,'' in \emph{IJCAI}, 2017.

\bibitem{Selvaraju_2019_ICCV}
R.~R. Selvaraju, S.~Lee, Y.~Shen, H.~Jin, S.~Ghosh, L.~Heck, D.~Batra, and D.~Parikh, ``Taking a hint: Leveraging explanations to make vision and language models more grounded,'' in \emph{Proceedings of the IEEE/CVF International Conference on Computer Vision (ICCV)}, October 2019.

\bibitem{selvaraju2017grad}
R.~R. Selvaraju, M.~Cogswell, A.~Das, R.~Vedantam, D.~Parikh, and D.~Batra, ``Grad-cam: Visual explanations from deep networks via gradient-based localization,'' in \emph{Proceedings of the IEEE international conference on computer vision}, 2017, pp. 618--626.

\bibitem{liu2021just}
\BIBentryALTinterwordspacing
E.~Z. Liu, B.~Haghgoo, A.~S. Chen, A.~Raghunathan, P.~W. Koh, S.~Sagawa, P.~Liang, and C.~Finn, ``Just train twice: Improving group robustness without training group information,'' in \emph{Proceedings of the 38th International Conference on Machine Learning}, ser. Proceedings of Machine Learning Research, M.~Meila and T.~Zhang, Eds., vol. 139.\hskip 1em plus 0.5em minus 0.4em\relax PMLR, 18--24 Jul 2021, pp. 6781--6792. [Online]. Available: \url{https://proceedings.mlr.press/v139/liu21f.html}
\BIBentrySTDinterwordspacing

\bibitem{hong2021unbiased}
\BIBentryALTinterwordspacing
Y.~Hong and E.~Yang, ``Unbiased classification through bias-contrastive and bias-balanced learning,'' 2021. [Online]. Available: \url{https://openreview.net/forum?id=2OqZZAqxnn}
\BIBentrySTDinterwordspacing

\bibitem{ji2019invariant}
X.~Ji, J.~F. Henriques, and A.~Vedaldi, ``Invariant information clustering for unsupervised image classification and segmentation,'' in \emph{Proceedings of the IEEE/CVF International Conference on Computer Vision}, 2019, pp. 9865--9874.

\bibitem{van2020scan}
W.~Van~Gansbeke, S.~Vandenhende, S.~Georgoulis, M.~Proesmans, and L.~Van~Gool, ``Scan: Learning to classify images without labels,'' in \emph{European Conference on Computer Vision}.\hskip 1em plus 0.5em minus 0.4em\relax Springer, 2020, pp. 268--285.

\bibitem{nam2022ssa}
\BIBentryALTinterwordspacing
J.~Nam, J.~Kim, J.~Lee, and J.~Shin, ``Spread spurious attribute: Improving worst-group accuracy with spurious attribute estimation,'' 4 2022. [Online]. Available: \url{https://arxiv.org/abs/2204.02070v1}
\BIBentrySTDinterwordspacing

\bibitem{zhang2018gce}
\BIBentryALTinterwordspacing
Z.~Zhang and M.~R. Sabuncu, ``Generalized cross entropy loss for training deep neural networks with noisy labels,'' \emph{Advances in Neural Information Processing Systems}, vol. 2018-December, pp. 8778--8788, 5 2018. [Online]. Available: \url{https://arxiv.org/abs/1805.07836v4}
\BIBentrySTDinterwordspacing

\bibitem{luo2022pseudo}
\BIBentryALTinterwordspacing
L.~Luo, D.~Xu, H.~Chen, T.-T. Wong, and P.-A. Heng, ``Pseudo bias-balanced learning for debiased chest x-ray classification,'' 3 2022. [Online]. Available: \url{https://arxiv.org/abs/2203.09860v1}
\BIBentrySTDinterwordspacing

\bibitem{zhang2023learning}
Y.-K. Zhang, Q.-W. Wang, D.-C. Zhan, and H.-J. Ye, ``Learning debiased representations via conditional attribute interpolation,'' in \emph{Proceedings of the IEEE/CVF Conference on Computer Vision and Pattern Recognition}, 2023, pp. 7599--7608.

\bibitem{zhang2024poisoning}
\BIBentryALTinterwordspacing
Y.~Zhang, Z.~Wang, R.~Hu, X.~Duan, Y.~ZHENG, B.~Huai, J.~Han, and J.~Sang, ``Poisoning for debiasing: Fair recognition via eliminating bias uncovered in data poisoning,'' in \emph{ACM Multimedia 2024}, 2024. [Online]. Available: \url{https://openreview.net/forum?id=jTtfDitRAt}
\BIBentrySTDinterwordspacing

\bibitem{Wang_2024_CVPR}
Y.~Wang, J.~Sun, C.~Wang, M.~Zhang, and M.~Yang, ``Navigate beyond shortcuts: Debiased learning through the lens of neural collapse,'' in \emph{Proceedings of the IEEE/CVF Conference on Computer Vision and Pattern Recognition (CVPR)}, June 2024, pp. 12\,322--12\,331.

\bibitem{papyan2020prevalence}
V.~Papyan, X.~Han, and D.~L. Donoho, ``Prevalence of neural collapse during the terminal phase of deep learning training,'' \emph{Proceedings of the National Academy of Sciences}, vol. 117, no.~40, pp. 24\,652--24\,663, 2020.

\bibitem{GoodBengCour16}
I.~J. Goodfellow, Y.~Bengio, and A.~Courville, \emph{Deep Learning}.\hskip 1em plus 0.5em minus 0.4em\relax Cambridge, MA, USA: MIT Press, 2016, \url{http://www.deeplearningbook.org}.

\bibitem{lloyd1982least}
S.~Lloyd, ``Least squares quantization in pcm,'' \emph{IEEE Transactions on Information Theory}, vol.~28, no.~2, pp. 129--137, 1982.

\bibitem{rousseeuw87silhouetteCluster}
\BIBentryALTinterwordspacing
P.~Rousseeuw, ``Silhouettes: a graphical aid to the interpretation and validation of cluster analysis,'' \emph{J. Comput. Appl. Math.}, vol.~20, no.~1, pp. 53--65, 1987. [Online]. Available: \url{http://portal.acm.org/citation.cfm?id=38772}
\BIBentrySTDinterwordspacing

\bibitem{lecun2010mnist}
Y.~LeCun, C.~Cortes, and C.~Burges, ``Mnist handwritten digit database,'' \emph{ATT Labs [Online]. Available: http://yann.lecun.com/exdb/mnist}, vol.~2, 2010.

\bibitem{liu2015faceattributes}
Z.~Liu, P.~Luo, X.~Wang, and X.~Tang, ``Deep learning face attributes in the wild,'' in \emph{Proceedings of International Conference on Computer Vision (ICCV)}, December 2015.

\bibitem{Rothe-IJCV-2018}
R.~Rothe, R.~Timofte, and L.~V. Gool, ``Deep expectation of real and apparent age from a single image without facial landmarks,'' \emph{International Journal of Computer Vision}, vol. 126, no. 2-4, pp. 144--157, 2018.

\bibitem{he2016deep}
K.~He, X.~Zhang, S.~Ren, and J.~Sun, ``Deep residual learning for image recognition,'' in \emph{Proceedings of the IEEE conference on computer vision and pattern recognition}, 2016, pp. 770--778.

\bibitem{eidinger2014age}
E.~Eidinger, R.~Enbar, and T.~Hassner, ``Age and gender estimation of unfiltered faces,'' \emph{IEEE Transactions on Information Forensics and Security}, vol.~9, no.~12, pp. 2170--2179, 2014.

\bibitem{Hendrycks2019}
\BIBentryALTinterwordspacing
D.~Hendrycks and T.~Dietterich, ``Benchmarking neural network robustness to common corruptions and perturbations,'' \emph{7th International Conference on Learning Representations, ICLR 2019}, 3 2019. [Online]. Available: \url{https://arxiv.org/abs/1903.12261v1}
\BIBentrySTDinterwordspacing

\bibitem{snoek2012practical}
J.~Snoek, H.~Larochelle, and R.~P. Adams, ``Practical bayesian optimization of machine learning algorithms,'' in \emph{Advances in neural information processing systems}, 2012, pp. 2951--2959.

\bibitem{wandb}
\BIBentryALTinterwordspacing
L.~Biewald, ``Experiment tracking with weights and biases,'' 2020, software available from wandb.com. [Online]. Available: \url{https://www.wandb.com/}
\BIBentrySTDinterwordspacing

\end{thebibliography}

\begin{IEEEbiography}[{\includegraphics[width=1in,height=1.25in,clip,keepaspectratio]{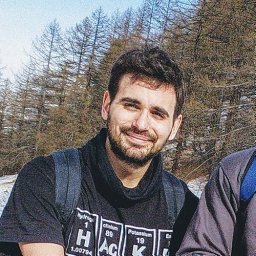}}]{Carlo Alberto Barbano}{\space}(Member, IEEE) is a postdoctoral researcher at University of Turin, Italy. He received the M.S. degree in computer science at University of Turin in 2020, cum laude. He became a Member of IEEE in 2022. In 2023, he received a double Ph.D. with honors at Institut Polytechnique de Paris (Télécom Paris), France and at University of Turin, Italy with the thesis ``Collateral-Free Learning of Deep Representations: From Natural Images to Biomedical Applications''. His research primarly focuses on representation learning, biomedical image analysis, and debiasing, with special interest in foundation models.
\end{IEEEbiography}

\begin{IEEEbiography}[{\includegraphics[width=1in,height=1.25in,clip,keepaspectratio]{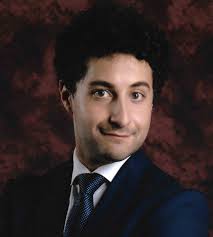}}]{Enzo Tartaglione}{\space}(Senior Member, IEEE) is Maître de Conférences at Télécom
Paris, and he is an Hi! Paris chair holder. He received the MS degree in Electronic Engineering at Politecnico di Torino in 2015, cum laude. The same year, he also received a magna cum laude MS in electrical and computer engineering at University of Illinois at Chicago. In 2016 he was also
awarded the MS in Electronics by Politecnico di Milano, cum laude. In 2019 he obtained the Ph.D. in Physics at Politecnico di Torino, cum laude, with the thesis ‘‘From Statistical Physics to Algorithms in Deep Neural Systems’’. His principal interests include compression, sparsification, pruning and watermarking of deep neural networks, deep learning for medical imaging, privacy-aware learning, data
debiasing, regularization for deep learning and neural networks growing. His expertise mainly focuses on the themes of efficient deep learning, with articles published on top conferences and journals in the field.
\end{IEEEbiography}

\begin{IEEEbiography}[{\includegraphics[width=1in,height=1.25in,clip,keepaspectratio]{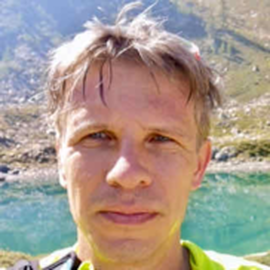}}]{Marco Grangetto}{\space}(Senior Member, IEEE) received the M.S.
degree in electrical engineering and the Ph.D. degree from
the Politecnico di Torino, Turin, Italy, in 1999 and 2003,
respectively. He is currently a Full Professor with the Department of Computer Science, Università di Torino, Turin,
where he coordinates research in the area of image processing and computer vision. His research interests are in the
fields of multimedia signal processing and networking. His
expertise includes wavelets, image and video coding, data
compression, video error concealment, error resilient video
coding, computer vision, and biomedical image processing.
Dr. Grangetto is a member of Moving Picture, Audio and
Data Coding by Artificial Intelligence (MPAI). He was the
recipient of the Premio Optime by the Unione Industriale
di Turin in September 2000 and a Fulbright Grant in 2001
for a research period with the Department of Electrical
and Computer Engineering, University of California, San
Diego, CA, USA. He participated in the ISO standardization activities on Part 11 of the JPEG2000 standard.
He was an Associate Editor of the IEEE TRANSACTIONS
ON COMMUNICATIONS and the IEEE TRANSACTIONS ON
MULTIMEDIA.
\end{IEEEbiography}

\appendix
\subsection{Experimental details}
\label{appendix:experiments-details}

In this section we provide the detailed description of the setup for all the experiments presented.
\newline

\subsubsection{Biased-MNIST}
We use the network architecture proposed by Bahng~\emph{et~al.}~\cite{bahng2019rebias}, consisting of four convolutional layers with $7 \times 7$ kernels. The EnD regularization term is applied on the average pooling layer, before the fully connected classifier of the network. 
Following Bahng~\emph{et~al.}, we use the Adam optimizer with a learning rate of $0.001$, a weight decay of $10^{-4}$ and a batch size of 256. 
We train for 80 epochs. We do not use any data augmentation scheme. We use 30\% of the training set as validation set, and we colorize it using a $\rho$ value of 0.1.
\newline

\subsubsection{CelebA}
Following Nam~\emph{et~al.}~\cite{nam2020learning}, we use the Adam optimizer with a learning rate of $0.001$, a batch size of 256, and a weight decay of $10^{-4}$. We train for 50 epochs. Images are resized to $224\times224$ and augmented with random horizontal flip. To construct the validation set, we sample $N$ images from each pair $(t, b)$ of the training set, where $N$ is 20\% the size of the least populated group $(t, b)$. 
The EnD hyperparameters $\alpha$ and $\beta$ are searched using the Bayesian optimization~\cite{snoek2012practical} implementation provided by \emph{Weights and Biases}~\cite{wandb} on the validation set, in the interval $[0;50]$.
To provide a mean performance along with the standard deviation, we select the top 3 models based on the best validation accuracy obtained, and we report the average accuracy on the final test sets.
\newline

\subsubsection{IMDB}
We use the Adam optimizer with a learning rate of 0.001, a batch size of 256 and a weight decay of $10^{-4}$. We train for 50 epochs. As with CelebA, images are resized to $224\times224$ and randomly flipped at training time for augmentation. In this case, it is not possible to construct a validation set including samples from both EB1 and EB2, without altering the test set composition. Hence, we perform a 4-fold cross validation for every experiment. For example, when training on EB1, we use one fold of EB2 as validation set and the remaining three folds as EB2 test set. We repeat this process until each EB2 fold is used both as validation and as test set. The same process is repeated when training on EB2, by splitting EB1 in validation and test folds. When training for age prediction, we follow Kim~\emph{et ~al.}~\cite{Kim_2019_CVPR}, by binning the age values in the intervals 0-19, 20-24, 25-29, 30-34, 34-39, 40-44, 45-49, 50-54, 55-59, 60-64, 65-69, 70-100, proposed by Alvi~\emph{et~al.}~\cite{alvi2018turning}.
For every fold, the EnD hyperparameters $\alpha$ and $\beta$ are searched using the Bayesian optimization~\cite{snoek2012practical} implementation provided by \emph{Weights and Biases}~\cite{wandb} on the validation set, in the interval $[0;50]$.
To provide a mean performance along with the standard deviation, we select the top model for each fold, based on the best validation accuracy obtained. We report the accuracy obtained on the final test sets, as average accuracy among the different folds. 

\subsection{Additional empirical results}
\label{sec:extra-additional-results}
In this section we provide some additional results about our debiasing technique, mainly focusing on the worst-case scenarios described in Section~\ref{sec:analysis-mnist}. 
\newline

\begin{table}
    \centering
    \begin{tabular}{c c c c}
        \toprule
        $\rho$ & Vanilla &  EnD (target) & EnD (random) \\
        \midrule
        0.995 & 72.10\std{1.90} & 72.25\std{0.56}  & 66.68\std{0.35} \\
        \bottomrule
    \end{tabular}
    \caption{Debiasing on incorrect bias labels. Target means that target labels are also used as bias labels (i.e. $t_i = b_i$, worst case), and random means that bias labels are assigned randomly.}
    \label{table:mnist-wrong-pseudo}
\end{table}

\subsubsection{Debiasing on an unbiased dataset}
\label{extra:debias-on-unbiased}

Here we show that the supervised EnD regularization does not deteriorate the final results if applied to a training set that is not biased. With $\rho=0.1$, applying the regularization term is not harmful to obtaining good generalization (Vanilla obtains 99.21\% and EnD achieves 99.24$\pm$0.05): this is because in a supervised setting we still have access to the correct color labels, thus we do not perform disentanglement over any useful features for the network.
This is a trivial result, however, with this demonstrated we can now focus on an unbiased training set in the unsupervised case.
\newline

\subsubsection{Debiasing with wrong pseudo-labels}
\label{extra:debias-on-wrong-bias}

\begin{figure}
    \centering
    \begin{subfigure}{0.45\columnwidth}
        \centering
        \includegraphics[width=1\columnwidth]{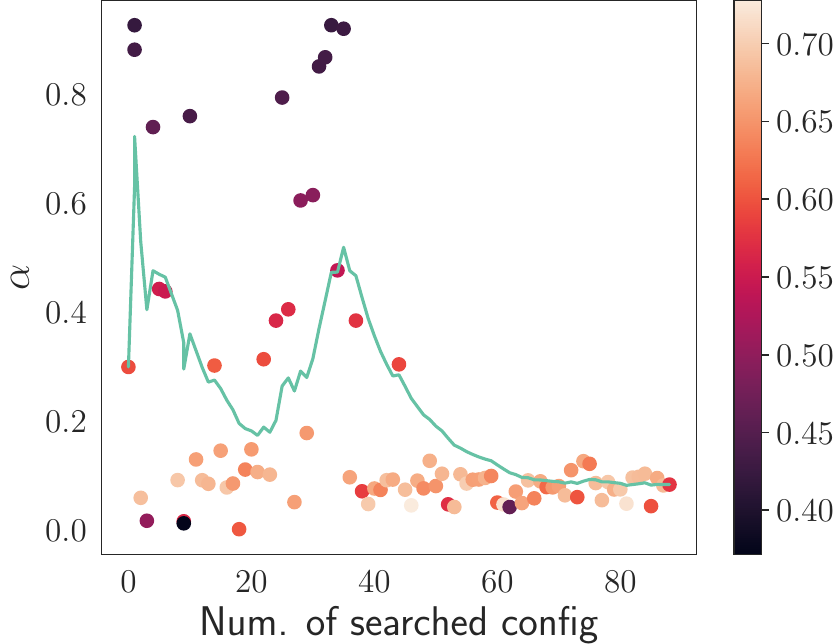}
        \caption{~}
        \label{fig:alpha-vs-time}
    \end{subfigure}%
    \hfill
    \begin{subfigure}{0.45\columnwidth}
        \centering
        \includegraphics[width=1\columnwidth]{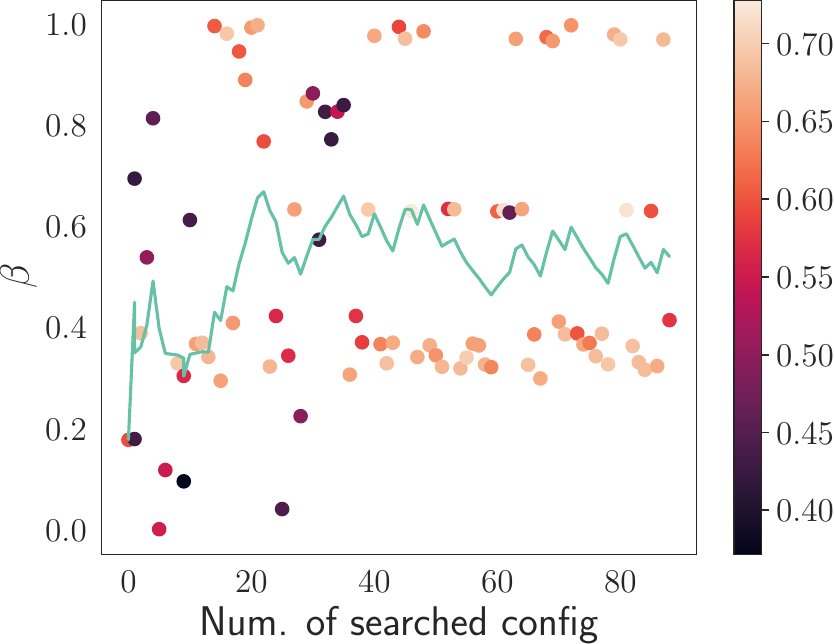}
        \caption{~}
        \label{fig:beta-vs-time}
    \end{subfigure}

    \caption{{Evolution of (a) $\alpha$ and (b) $\beta$ versus the number of searched configs} during the hyperparameters optimization with incorrect pseudo-labels. We can observe how the optimization process drives $\alpha$ towards 0, while $\beta$ does not seem to be relevant. The point color indicates the accuracy of the unbiased test set, while the line shows the trend as an exponentially weighted moving average computed with a smoothing factor of $0.1$.}
    \label{fig:alpha-beta-vs-time}
\end{figure}

We now assume that the pseudo-labels we compute are not representative of the true bias attributes. 
Using Biased-MNIST as a case study, we identify that the worst-case scenario for the pseudo-labeling step corresponds to using a completely unbiased dataset (i.e. $\rho=0.1$) for training the biased encoder.
Taking into account the results shown in Section~\ref{extra:debias-on-unbiased}, performing the pseudo-labeling step in this setting will most likely result in pseudo-labels corresponding to the actual target class rather than the background color.
We emulate this event by setting the bias label $b_i$ equal to the target label $t_i$ for every sample in the dataset, and then we the apply EnD algorithm. 
To test this worst-case with EnD, we choose $\rho=0.995$ as it provides a way for the final accuracy to both decrease or increase with respect to a vanilla model. 
The results are reported in Table~\ref{table:mnist-wrong-pseudo} and noted as \emph{target}. Even in this case, we are able to retain the baseline performances, although we do not obtain any significant improvement. 
This is thanks to the hyperparameter optimization policy that we employ (recall that we assume an unbiased validation set - even if small - is available). 
Figure~\ref{fig:alpha-beta-vs-time} visualizes the evolution of the hyperparameters $\alpha$ and $\beta$ while searching for possible configurations. 
In this setting, $\alpha$ represents the most dangerous term, as it enforces decorrelation among samples with the same class, conflicting with the cross-entropy term. 
However, the optimization process drivers $\alpha$ towards 0, making it effectively non-influent on the loss term. 
On the other hand, the entangling term $\beta$ does not bring any contribution to the learning process: it is, in fact, useless as there is full alignment between target and bias labels, hence $B(i)$ = \O. 
A possible scenario in which $\beta$ would not have null influence is if we do not impose $t_i = b_i$. 
We explore the extreme setting by assigning a random value to $b_i$ for every sample $i$. The results are reported as \emph{random} in Table~\ref{table:mnist-wrong-pseudo}. In this case, it is possible to observe a drop in performance with respect to the baseline. However, we argue that random pseudo-labels would be the result of poor representations due to possibly underfitting models or lack of sufficient training data - which, in a practical setting, would be a more pressing issue.

\subsection{Complete derivations for Section~\ref{sec:bias-theoretical-model}}
\label{sec:appendix-theoretical-derivations}
In this section we present the full derivations of the theoretical results.
\newline

\subsubsection{Derivation of Eq.\eqref{eq:MI_perfect}}
\label{appendix:MI_perfect}
Let \eqref{eq:entropy_assumption}, we can write the conditional entropy
\begin{align}
    H_{perf}&(B|Y) = -\sum_{i=1}^{N_T} P(y_i) \sum_{j=1}^{N_T} P(b_j | y_i) \log_{2} \left [P(b_j | y_i) \right ]\nonumber\\
    =& -\frac{1}{N_T} \sum_{i=1}^{N_T} \sum_{j=1}^{N_T} P(b_j | y_i) \log_{2} [P(b_j | y_i)]\nonumber\\
    =& -\frac{1}{N_T}  \sum_{i=1}^{N_T} \sum_{j=1}^{N_T} \left\{ [\rho \log_{2}(\rho)] \delta_{ij} +\right .\nonumber\\
    &\left . +\left[\frac{1-\rho}{N_T-1} \log_{2}\left(\frac{1-\rho}{N_T-1}\right)\right]\overline{\delta}_{ij} \right\}\nonumber\\
    =& -\frac{1}{N_T} \left\{ N_T\cdot [\rho \log_{2}(\rho)] +(N_T^2 - N_T)\cdot \right .\nonumber\\
    &\left . \cdot\left[\frac{1-\rho}{N_T-1}\cdot \log_{2}\left(\frac{1-\rho}{N_T - 1}\right)\right]\right\}\nonumber
\end{align}
\begin{align}
    =& - \rho \cdot \log_{2}(\rho) - \left\{(1-\rho)\cdot [\log_{2}(1-\rho) - \log_{2}(N_T-1)]\right\}\nonumber\\
    =& - \rho \cdot \log_{2}(\rho) - \log_{2}(1-\rho) +\log_{2}(N_T-1) +\nonumber\\ 
    & +\rho \cdot \log_{2}(1-\rho) - \rho \cdot\log_{2}(N_T-1)\nonumber\\
    =& \rho \cdot \log_{2}\left[\frac{1-\rho}{(N_T-1) \cdot\rho}\right] + \log_{2}\left[\frac{N_T-1}{1-\rho}\right] \nonumber\\
    =&\log_{2}\left[\frac{1-\rho}{N_T-1}\right]^{\rho-1} + \rho \log_2 \rho\nonumber\\
    =& -\log_{2}\left[\frac{1-\rho}{N_T-1}\right]^{1-\rho} + \log_2 \rho\cdot \rho^{\rho-1}\nonumber\\
    =& -\log_{2}\rho\left[\frac{1-\rho}{\rho(N_T-1)}\right]^{1-\rho}
\end{align}

Once we have the conditional entropy, we can compute the mutual information

\begin{align}
    I_{perf}(B, Y) &= H(B) - H_{perf}(B|Y)\nonumber\\ 
    &= \log_2 N_T +\log_{2}\rho\left[\frac{1-\rho}{\rho(N_T-1)}\right]^{1-\rho}
\end{align}

from which, exploiting \eqref{eq:entropy_assumption}, we can obtained the normalized mutual information
\begin{align}
    \hat{I}_{perf}(B,Y) =& \frac{2\cdot I_{perf}(B, Y)}{H(B) + H(Y)}\nonumber\\
    &=\frac{2\cdot\left\{ \log_2 N_T +\log_{2}\rho\left[\frac{1-\rho}{\rho(N_T-1)}\right]^{1-\rho}\right\}}{2\cdot \log_2 N_T}\nonumber\\
    &= 1 + \log_{N_T}\left\{\rho\left[\frac{1-\rho}{\rho(N_T-1)}\right]^{1-\rho}\right\}\nonumber\\ 
    &= \log_{N_T}\left\{N_T\rho\left[\frac{1-\rho}{\rho(N_T-1)}\right]^{1-\rho}\right\}\nonumber
\end{align}
\newline

\subsubsection{Derivation of \eqref{eq:MIcool}}
\label{appendix:MI}
Let the marginal as in \eqref{eq:joint_bz}. Following the definition of mutual information, we can write
\begin{align*}
    I&(B,Y) = \sum_{i,j} P(b_j,y_i) \log_2\frac{P(b_j,y_i)}{P(b_j)P(y_i)}\nonumber\\
    =&\frac{1}{N_T} \left\{ N_T \cdot\left[\rho (1 - \varepsilon) + \phi(1-\rho) \right] \cdot \right. \nonumber\\
    &\cdot \log_2 (N_T \cdot(\rho (1 - \varepsilon) + \phi(1-\rho))) +\nonumber \\
    &+ (N_T^2 - N_T) \cdot \left [\frac{(1-\phi)(1-\rho)}{N_T-1} + \frac{\rho\varepsilon}{N_T-2+\rho}\right ]\cdot\nonumber\\
    &\left .\cdot \log_2 \left[ N_T \cdot \left(\frac{(1-\phi)(1-\rho)}{N_T-1} + \frac{\rho\varepsilon}{N_T-2+\rho}\right)\right ] \right\}\nonumber
\end{align*}
\begin{align*}
    =&\left[\rho (1 - \varepsilon) + \phi(1-\rho) \right] \cdot \nonumber\\
    &\cdot \left[\log_2 N_T + \log_2(\rho (1 - \varepsilon) + \phi(1-\rho)\right] \nonumber\\
    &+ \left [(1-\phi)(1-\rho) + \frac{(N_T - 1)\rho\varepsilon}{N_T-2+\rho}\right ]\cdot\nonumber\\
    &\cdot \left[ \log_2 N_T + \log_2 \left(\frac{(1-\phi)(1-\rho)}{N_T-1} + \frac{\rho\varepsilon}{N_T-2+\rho}\right)\right ]\nonumber
\end{align*}
\begin{align}
    =&\log_2\left[\rho (1 - \varepsilon) + \phi(1-\rho)\right]^{\rho (1 - \varepsilon) + \phi(1-\rho)}\nonumber\\
    &+ \log_2 \left(\frac{(1-\phi)(1-\rho)}{N_T-1} + \frac{\rho\varepsilon}{N_T-2+\rho}\right)^{(1-\phi)(1-\rho) + \frac{(N_T - 1)\rho\varepsilon}{N_T-2+\rho}}\nonumber\\
    &+\log_2 N_T \left[\rho (1 - \varepsilon) + \phi(1-\rho) +\right.\nonumber\\
    &+\left .(1-\phi)(1-\rho) + \frac{(N_T - 1)\rho\varepsilon}{N_T-2+\rho}\right ]\nonumber\\
    =&\log_2\left[\rho (1 - \varepsilon) + \phi(1-\rho)\right]^{\rho (1 - \varepsilon) + \phi(1-\rho)} \nonumber\\
    &+ \log_2 \left(\frac{(1-\phi)(1-\rho)}{N_T-1} + \frac{\rho\varepsilon}{N_T-2+\rho}\right)^{(1-\phi)(1-\rho) + \frac{(N_T - 1)\rho\varepsilon}{N_T-2+\rho}}\nonumber\\
    &+\log_2 N_T \left[1-\rho\varepsilon + \frac{(N_T - 1)\rho\varepsilon}{N_T-2+\rho}\right ]\label{eq:IBZA}
\end{align}
Similarly to how been done in Sec.\ref{appendix:MI_perfect}, we can obtain the normalized mutual information
\begin{align}
    \hat{I}&(B,Y) = \frac{2\cdot I(B, Y)}{H(B) + H(Y)} = \frac{I(B, Y)}{\log_2 N_T}\nonumber\\
    =& \log_{N_T}\left[\rho (1 - \varepsilon) + \phi(1-\rho)\right]^{\rho (1 - \varepsilon)+ \phi(1-\rho)}\nonumber\\
    &+ \log_{N_T} \left [\frac{(1-\phi)(1-\rho)}{N_T-1} + \frac{\rho\varepsilon}{N_T-2+\rho}\right ]^{(1-\phi)(1-\rho) + \frac{(N_T-1)\rho\varepsilon}{N_T-2+\rho}}\nonumber\\
    &+ 1 - \rho\varepsilon + \frac{(N_T-1)\rho\varepsilon}{N_T-2-\rho}\nonumber
\end{align}
which in this case results in a simple change of base for the logarithms in \eqref{eq:IBZA}.
\newline

\subsubsection{Derivation of biasedness/unfairness}
\label{appendix:v-derivation}
Using the theoretical model presented in Section~\ref{sec:theoretical-model}, we can compute $\phi$ by rewriting of Eq.~\ref{eq:joint_bz} in order to derive it. First, for easier readibility, we explicitly enumerate the cases given by the kronecker deltas of Eq.~\ref{eq:joint_bz}:
\begin{equation}
    P(b, y) = 
\begin{cases}
    \frac{1}{N_T} \left[\rho (1 - \varepsilon) + \phi(1-\rho) \right] \quad \text{if } b=y \\[2ex]
	\frac{1}{N_T}\left [\frac{(1-\phi)(1-\rho)}{N_T-1} + \frac{\rho\varepsilon}{N_T-2+\rho} \right]  \quad \text{if } b \neq y
\end{cases}
\end{equation}
From this, we can rewrite each case in order to obtain the bias tendency $\phi_{b,y}$ for each $b$ and $y$:
\begin{equation}
    \phi_{b,y} = 
\begin{cases}
     P(b, y)\frac{N_T}{1 - \rho} - \frac{\rho(1-\varepsilon)}{1 - \rho} \quad \text{if } b = y \\[2ex]
     1 - P(b,y)\frac{N_T^2-N_T}{1-\rho} + \varepsilon\frac{\rho(N_T-1)}{(1-\rho)(N_T-2-\rho)} \quad \text{if } b \neq y
\end{cases}
\end{equation}
Note that, by doing so, we are actually computing multiple $\phi$ values, as a function of $b$ and $y$, while in Eq.~\ref{eq:joint_bz} we assume the values to be constant (i.e. the bias tendency is the same independently from the value of $b$ and $y$ considered). 
In our experiments, to account for the assumptions we make ($\varepsilon = 0$) and for measurements errors due to the stochastic nature of the training, we clip the measured joint probability between $[\frac{\rho}{N_T}; \frac{1}{N_T}]$ for the diagonal values ($b=y$), and between $[0; \frac{1-\rho}{N_T(N_T-1)}]$ for the off-diagonal values ($b \neq y$). This ensures that the computed $\phi_{b,y}$ lies between the valid range $[0; 1]$.
Finally, we can compute the global $\phi$ by averaging all the $\phi_{b,y}$:
\begin{equation}
\phi = \frac{1}{N_B N_T}\sum_{b,y} \phi_{b,y}
\end{equation}
where $N_B$ is the number of bias classes.

\end{document}